\documentclass[a4paper,fleqn]{cas-dc}

\usepackage{amsmath} 
\usepackage{bm} 

\usepackage{diagbox}
\usepackage{slashbox}
\usepackage{multirow}
\usepackage[numbers]{natbib}
\usepackage{booktabs}
\usepackage{graphicx}
\usepackage{array}
\usepackage{tabularx}
\usepackage{soul} 
\usepackage{xcolor}
\usepackage{threeparttable}
\usepackage{siunitx}
\usepackage{flushend}
\sethlcolor{yellow} 

\def\tsc#1{\csdef{#1}{\textsc{\lowercase{#1}}\xspace}}
\tsc{WGM}
\tsc{QE}
\tsc{EP}
\tsc{PMS}
\tsc{BEC}
\tsc{DE}

\begin{document}
\let\WriteBookmarks\relax
\def\floatpagepagefraction{1}
\def\textpagefraction{.001}
\shorttitle{\textit{Robotics and Autonomous Systems}}
\shortauthors{\textit{T. Okawara et~al.}}

\title [mode = title]{Tightly-Coupled LiDAR-IMU-Wheel Odometry with an Online Neural Kinematic Model Learning via Factor Graph Optimization}

\tnotemark[1]

\tnotetext[1]{This work was supported in part by JSPS KAKENHI Grant Number 22KJ0292 and a project commissioned by the New Energy and Industrial Technology Development Organization (NEDO).}


\author[1]{Taku Okawara}
\fnmark[*]
\ead{okawara.taku.t3@dc.tohoku.ac.jp}
\credit{Conceptualization of this study, Methodology, Software}

\affiliation[1]{organization={The Space Robotics Lab. in the Department of Aerospace Engineering, Graduate School of Engineering, Tohoku University},
                city={Sendai, Miyagi},
                country={Japan}}

\author[2]{Kenji Koide}
\author[2]{Shuji Oishi}
\author[2]{Masashi Yokozuka}
\author[2]{Atsuhiko Banno}
\author[1]{Kentaro Uno}
\author[1]{Kazuya Yoshida}


\affiliation[2]{organization={National Institute of Advanced Industrial Science and Technology (AIST)},
                city={Tsukuba, Ibaraki},
                country={Japan}}

\cortext[cor1]{Corresponding author.}


\begin{abstract}
Environments lacking geometric features (e.g., tunnels and long straight corridors) are challenging for LiDAR-based odometry algorithms because LiDAR point clouds degenerate in such environments.
For wheeled robots, a wheel kinematic model (i.e., wheel odometry) can improve the reliability of the odometry estimation.  
However, the kinematic model suffers from complex motions (e.g., wheel slippage, lateral movement) in the case of skid-steering robots particularly because this robot model rotates by skidding its wheels.
Furthermore, these errors change nonlinearly when the wheel slippage is large (e.g., drifting) and are subject to terrain-dependent parameters.
To simultaneously tackle point cloud degeneration and the kinematic model errors, we developed a LiDAR-IMU-wheel odometry algorithm incorporating online training of a neural network that learns the kinematic model of wheeled robots with nonlinearity.
We propose to train the neural network online on a factor graph along with robot states, allowing the learning-based kinematic model to adapt to the current terrain condition.
The proposed method jointly solves online training of the neural network and LiDAR-IMU-wheel odometry on a unified factor graph to retain the consistency of all those constraints.
Through experiments, we first verified that the proposed network adapted to a changing environment, resulting in an accurate odometry estimation across different environments.
We then confirmed that the proposed odometry estimation algorithm was robust against point cloud degeneration and nonlinearity (e.g., large wheel slippage by drifting) of the kinematic model.
The summary video is available here: \textcolor{blue}{\url{https://www.youtube.com/watch?v=CvRVhdda7Cw}}
\end{abstract}



\begin{keywords}
  \sep  Training neural network on factor graph \sep Odometry \sep Sensor fusion \sep Machine learning \sep Factor graph optimization \sep Calibration \sep Skid-steered robot
\end{keywords}

\maketitle
\section{Introduction}
Accurate and robust odometry estimation is crucial for autonomous robots to accomplish reliable navigation.
State-of-the-art odometry algorithms based on Light Detection and Ranging--Inertial Measurement Unit (LiDAR-IMU) odometry algorithms~\cite{xu2022fast,qin2020lins,shan2020lio} accurately estimate the robot pose owing to the tight fusion of LiDAR and IMU data.
The IMU-based constraint can make the odometry estimation more robust against rapid motion and short-term point cloud degeneration.
However, it is still difficult for LiDAR-IMU odometry algorithms to overcome long-term point cloud
degeneration in featureless environments (e.g., tunnels and long corridors). IMU-based constraints cannot avoid an accumulation of errors that turn into estimation drift and corruption in such challenging environments.

A wheel encoder can provide a reliable constraint in such environments where LiDAR point clouds degenerate because the wheel odometry estimation can provide accurate motion prediction compared with an IMU owing to integral errors.
While double integration of the acceleration is needed for calculating translational displacement in the case of IMU-based odometry estimation, wheel encoders measure wheel angular velocities and need only a single integration that accumulates errors more slowly.
Therefore, LiDAR-IMU-wheel odometry has the potential to overcome long-term LiDAR point cloud degeneration. 
Although the differential drive model is often used as a kinematic model of wheeled robots due to its simplicity, this model ignores both lateral movement and wheel slippage; thus, such complex motions cannot be accurately expressed with this simple model.
Incorporating both lateral movement and wheel slippage can reduce the drift; in the case of skid-steering robots particularly, lateral movement and wheel slippage have a large impact on the estimation accuracy~\cite{anousaki2004dead} because this robot model rotates by skidding its wheels based on different angular velocities of the left and right wheels.
Furthermore, wheel slippage depends on terrain conditions, and thus the wheel odometry model must be maintained online to adapt to each environment.
Therefore, online calibration of the complex kinematic model is crucial for creating reliable wheel odometry-based constraints.

To estimate an accurate kinematic model of a skid-steering robot, our previous work~\cite{okawara2024tightly} conducted online calibration of the kinematic model based on the \textit{full linear model}~\cite{anousaki2004dead}.
Specifically, this work jointly solved the online calibration problem and tightly coupled LiDAR-IMU-wheel odometry such that the calibrated model-based robot motion, LiDAR-based motion, and IMU-based motion become consistent.
However, this linear model cannot express nonlinearity in the robot model (e.g., large wheel slippage during high-speed operation~\cite{brach2011tire, reichensdorfer2018stability}), because the robot motion model is regarded as linear.
A natural approach to overcome the errors caused by non-linearity is to introduce a robot motion model based on a neural network.
However, while the rigorous wheel odometry model depends on terrain-dependent parameters (e.g., friction coefficients and slip ratio), an offline trained network cannot express these parameters to adapt to changes in terrain conditions~\cite{onyekpe2021whonet}.

In this study, we propose a tightly coupled LiDAR-IMU-wheel odometry algorithm that incorporates online training of the neural network inferencing the kinematic model of a skid-steering robot.
Through online training, the proposed network can adapt to the current terrain condition while achieving accurate motion prediction owing to its non-linearity expression capability.
We jointly solve the online training of the network and LiDAR-IMU-wheel odometry on a unified factor graph in a tightly coupled way.

In this work, we focused on applying the proposed method to a skid-steering robot because identifying the complex model of this robot type is difficult and valuable; however, the proposed method can be applied to other robot platforms to accurately capture the nonlinearity of each kinematic model.

We extend our previous work~\cite{okawara2024tightly}, which jointly solves odometry estimation and online calibration of the full linear model with the following contributions:
\begin{enumerate}
  \item To deal with severe conditions, such as point cloud degeneration and nonlinear kinematic behavior of a skid-steering robot, we proposed a tightly coupled LiDAR-IMU-wheel odometry algorithm incorporating online training of the neural network to describe the kinematic model of a skid-steering robot. The neural network can implicitly express the nonlinearity of the kinematic model; thus, this learning-based motion constraint can compliantly adapt to the current terrain condition.
  \item To strike a balance between the online training speed and estimation accuracy, we designed the network to incorporate both an \textit{online learning model} and \textit{offline learning model}.
  While the \textit{offline learning model} expresses fixed features (i.e., terrain-independent terms), the \textit{online learning model} describes dynamically changing features (i.e., terrain-dependent terms).
  \item We proposed \textit{neural adaptive odometry factor} to directly optimize the \textit{online learning model} of the network on a factor graph such that the online learning-based motion constraint and LiDAR-IMU-wheel odometry become consistent.
\end{enumerate}

\section{Related work}\label{related_works}

\subsection{LiDAR-IMU odometry}\label{subsec:LIO}
The state-of-the-art LiDAR-IMU odometry algorithms enable accurate odometry estimation owing to a tight coupling of the LiDAR and IMU constraints. 
While a loose coupling approach conducts optimization based on poses estimated by measurements from each sensor (e.g., LiDAR and IMU), the tight coupling approach processes each sensor's raw measurement data to optimize unified constraints defined by each sensor's data.
This approach can estimate odometry accurately and robustly because the data from both sensors are directly fused.

LINS~\cite{qin2020lins} was the first tightly coupled LIDAR-IMU odometry algorithm with an iterated error-state Kalman filter.
This approach fused the IMU-based motion model and point cloud matching to update states iteratively.
LINS iteratively updates point correspondences and the sensor state until the state converges.
FAST-LIO2~\cite{xu2022fast} also uses an iterated Kalman filter for a tight fusion of LiDAR and IMU data.
Specifically, this work directly registered raw points of a new scan to a large local map (e.g.,~\si{1}{km}) due to the ikd-tree, which efficiently represents a dense and large map because of efficient nearest search and incremental updates of the structure. 
This registration method and the ikd-tree enabled fast, robust, and accurate odometry estimation, and thus FAST-LIO2 exhibited significantly higher accuracy compared to LINS.

Tightly coupled LiDAR-IMU odometry can estimate accurate robot poses in feature-rich environments where point cloud matching works well.
However, point cloud-based constraints become corrupted in featureless environments where point clouds degenerate for long periods (e.g., tunnels).
IMU-based constraints also face accumulation errors related to integration errors (particularly for accelerometers).
Therefore, a challenge of LiDAR-IMU odometry is robustness to point cloud degeneration.

\subsection{Learning-based odometry using proprioceptive sensors}\label{subsec:learning_based_calibration}
To improve robustness in the featureless environments described in Section~\ref{subsec:LIO}, proprioceptive sensors (e.g., wheel encoders and IMUs) are commonly used along with LiDARs because these sensors' data are independent of the geometric features of environments.
We can simply calculate relative displacement by integrating a kinematic model formulated by the proprioceptive sensor data.
However, this integration suffers from kinematic model error, bias and noise of the sensor measurements, and thus the odometry estimation has a large drift.
Therefore, many researchers have proposed a learning-based approach to implicitly express an accurate kinematic model for reducing the drift.

IONet~\cite{chen2018ionet} demonstrated an accurate three-Degrees-of-Freedom (3DoF) odometry estimation due to time-dependent features of IMU measurements and body motions based on the Long Short-Term Memory (LSTM) network.
IONet outperformed the traditional model-based method as in strap-down inertial navigation systems (SINS)~\cite{savage1998strapdown}.
In addition, TLIO~\cite{liu2020tlio} accurately estimated sensor inertial states, including IMU pose, velocity, and bias, with only IMU measurements through a tight coupling of a neural network-based 3D displacement estimation and a model-based kinematic model with a Kalman filter.

Onyekpe et~al. proposed a Recurrent Neural Network (RNN)-based wheel odometry model named WhONet~\cite{onyekpe2021whonet}. WhONet tackled challenges~\cite{onyekpe2021learning} of wheel odometry estimation such as wheel slippage, sharp cornering, and acceleration changes due to the RNN's ability to capture relationships within sequential sensor data.
This work suggested that the difficulty stems from a model adaptation to changes in terrain conditions and robot platform because WhONet was trained offline.

The kinematic model of skid-steering robots heavily depends on terrain-dependent parameters (e.g., friction coefficients and slip ratio), and thus online learning is desirable for obtaining an accurate kinematic model.
Navone et~al. conducted online training for the kinematic model of a skid-steering robot~\cite{navone2023online} with IMU and wheel encoders.
This work showed that the online learning-based odometry estimation outperformed the traditional extended Kalman filter approach because complex nonlinearity is implicitly expressed.
This online learning model had almost the same accuracy as the batch-learning model when the robot moved over the same terrain.
Furthermore, Ordonez et~al. trained online the kinematic and dynamic model of a skid-steering robot for expressing terrain-dependent terms~\cite{ordonez2017learning}. 
This work validated accurate wheel odometry estimation even in different terrain conditions owing to online learning, which enables the model to adapt to each environment.
These approaches only conducted online learning of the kinematic model based on reference data (e.g., visual odometry) obtained by exteroceptive sensors.
Therefore, these approaches did not fuse the online trained kinematic model and exteroceptive sensors.

\begin{figure}[tb]
  \centering
  \includegraphics[width=1\linewidth]{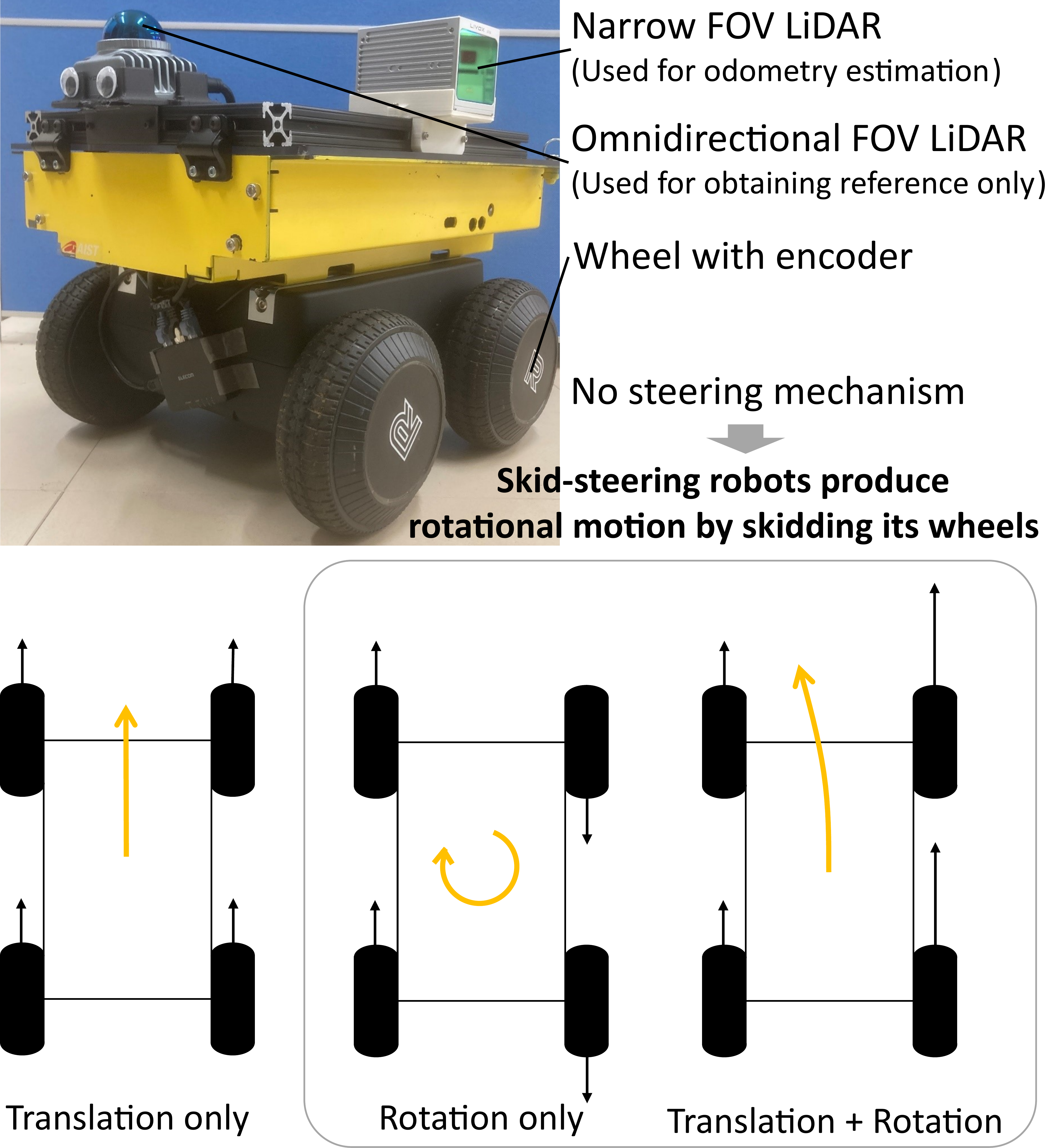}
  \caption{Skid-steering robot used as a testbed and diagram of motion of this robot model.
  Skid-steering robots produce rotational motion by skidding their wheels based on different angular velocities of the left and right wheels, instead of being equipped with a steering mechanism.
  We must consider the wheel slippage and lateral motion to obtain an accurate kinematic model of a skid-steering robot.}
  \label{fig:skid_steering_robot}
\end{figure}

\subsection{Model-based kinematic model for skid-steering robots}\label{subsec:model_based_offline_calibration}
As seen in Figure~\ref{fig:skid_steering_robot}, a skid-steering robot does not have any steering mechanism, and thus rotational motion is produced by skidding its wheels based on different angular velocities between the left and right wheels.
This robot model can be used in various fields due to its high road ability and mechanical simplicity.
However, this mechanical simplicity leads to the difficulty of modeling the kinematics accurately because wheel slippage and lateral motions must be considered in the model.
Due to the complexity of the wheel odometry model, various modeling approaches have been proposed~\cite{anousaki2004dead,le1997estimation,mandow2007experimental,wang2015analysis,rabiee2019friction}.

The full linear model~\cite{anousaki2004dead} is a generic model in that the velocity of the robot is assumed to be proportional to the angular velocity of the wheels.
This model does not require explicitly expressing the kinematic model, which is different from the following approaches~\cite{mandow2007experimental, wang2015analysis}.
Mandow et~al. proposed an extended differential drive model that introduced Instantaneous Centers of Rotation (ICR) parameters to better model the motion of skid-steering robots by extending the ideal differential drive model~\cite{mandow2007experimental}.
Wang et~al. also introduced a Radius Of Curvature (ROC)-based model that captures the relationship between the ICR parameters and ROC of the robot motion~\cite{wang2015analysis}.

Baril et~al. evaluated the above kinematic models in different environments, such as a flat plane and uneven terrain.
This work showed that the full linear model~\cite{anousaki2004dead} is the most accurate among these models because it has the flexibility to tolerate model errors that result from the difficulty of rigorous expression~\cite{baril2020evaluation}.
However, this work also suggested that the full linear model cannot accurately predict nonlinear motion in rough terrain.

\subsection{Proprioceptive and exteroceptive sensor fusion-based odometry for wheeled robots}\label{subsec:sensor_fusion__calibration}
Fusing exteroceptive sensors (e.g., LiDARs and cameras) with proprioceptive sensors (e.g., wheel encoders and IMUs) enables robust odometry estimation by complementing each sensor's properties~\mbox{\cite{yuan2023liwo}}.
However, accuracy of the wheel odometry model suffers from errors in kinematic parameters (e.g., wheel radius) and terrain-dependent parameters (e.g., friction coefficients and slip ratio).
Thus, many approaches corrected these errors dynamically based on online calibration by incorporating exteroceptive sensors.
In contrast to works solving sensor fusion-based odometry estimation solely~\mbox{\cite{yuan2023liwo}}, Simultaneous Calibration, Localization, and Mapping~\cite{kummerle2011simultaneous, kummerle2011simultaneous_journal} solves LiDAR-based SLAM and online calibration problems jointly.
Specifically, these works calibrated kinematic parameters (wheel radii and wheelbase of the differential drive robot) and extrinsic parameters (2D relative pose between the LiDAR frame and the robot frame) such that wheel odometry-based and LiDAR-based motions become consistent.
Extending these works, Lee et~al. fused camera, IMU, and wheel encoder data to solve SLAM and online calibration jointly for correcting a time offset between IMU and encoder data in addition to kinematic parameters and extrinsic parameters~\cite{lee2020visual}.
Murai et~al. conducted distributed simultaneous localization and online calibration of extrinsic parameters via Gaussian belief propagation in a factor graph~\cite{murai2024distributed}.

The aforementioned online calibration methods aim to correct the kinematic model of a differential drive robot under the ideal condition that wheel slippage and lateral motion are not present.
Some approaches conducted online calibration for a kinematic model of skid-steering robots~\cite{zuo2019visual, zuo2022visual, okawara2024tightly}.
Zuo et~al. applied an ICR-based model~\cite{mandow2007experimental} to the kinematic model of a skid-steering robot, and jointly solved online calibration of its model and visual-IMU SLAM~\cite{zuo2019visual,zuo2022visual}.
These works enabled a more accurate estimation than the visual-IMU fusion owing to the constraint of wheel odometry incorporating online calibration.
Our previous work conducted online calibration of the full linear model~\cite{anousaki2004dead}, which can implicitly express unknown kinematic parameters and environment-dependent values that are difficult to model rigorously~\cite{okawara2024tightly}.
In contrast to~\cite{zuo2019visual,zuo2022visual}, our previous work demonstrated accurate odometry estimation even in a featureless environment (e.g., long corridors) by handling degeneracy of LiDAR point clouds.

\subsection{Training neural networks on a factor graph}\label{subsec:training_on_factor_graph}
The online calibration methods described in Section~\ref{subsec:sensor_fusion__calibration} assume a linear model for skid-steering robots, and thus accuracy is not ensured for nonlinearity.
The nonlinearity can be implicitly expressed by a neural network, and thus we extended our previous work~\cite{okawara2024tightly} by replacing the linear model-based constraint with a neural network-based one.
Some odometry algorithms~\cite{buchanan2022learning, van2022learning} fused a neural network-based motion model and other constraints by factor graph optimization; however, the network was trained offline independent of odometry estimation.
The neural networks constrained robot poses based on fixed models and were not directly and dynamically trained on the factor graph.
Different from the aforementioned works, the proposed method jointly accomplishes tightly coupled LiDAR-IMU-wheel odometry and online training of the kinematic model for skid-steering robots.
The proposed method trains a network in a factor graph such that the neural network-based constraint and all other constraints (i.e., LiDAR-IMU-wheel odometry) are consistent.

 
\section{Methodlogy}\label{Methodlogy}

\subsection{System overview}\label{Overview}
\begin{figure*}[tb]
  \centering
  \includegraphics[width=1\linewidth]{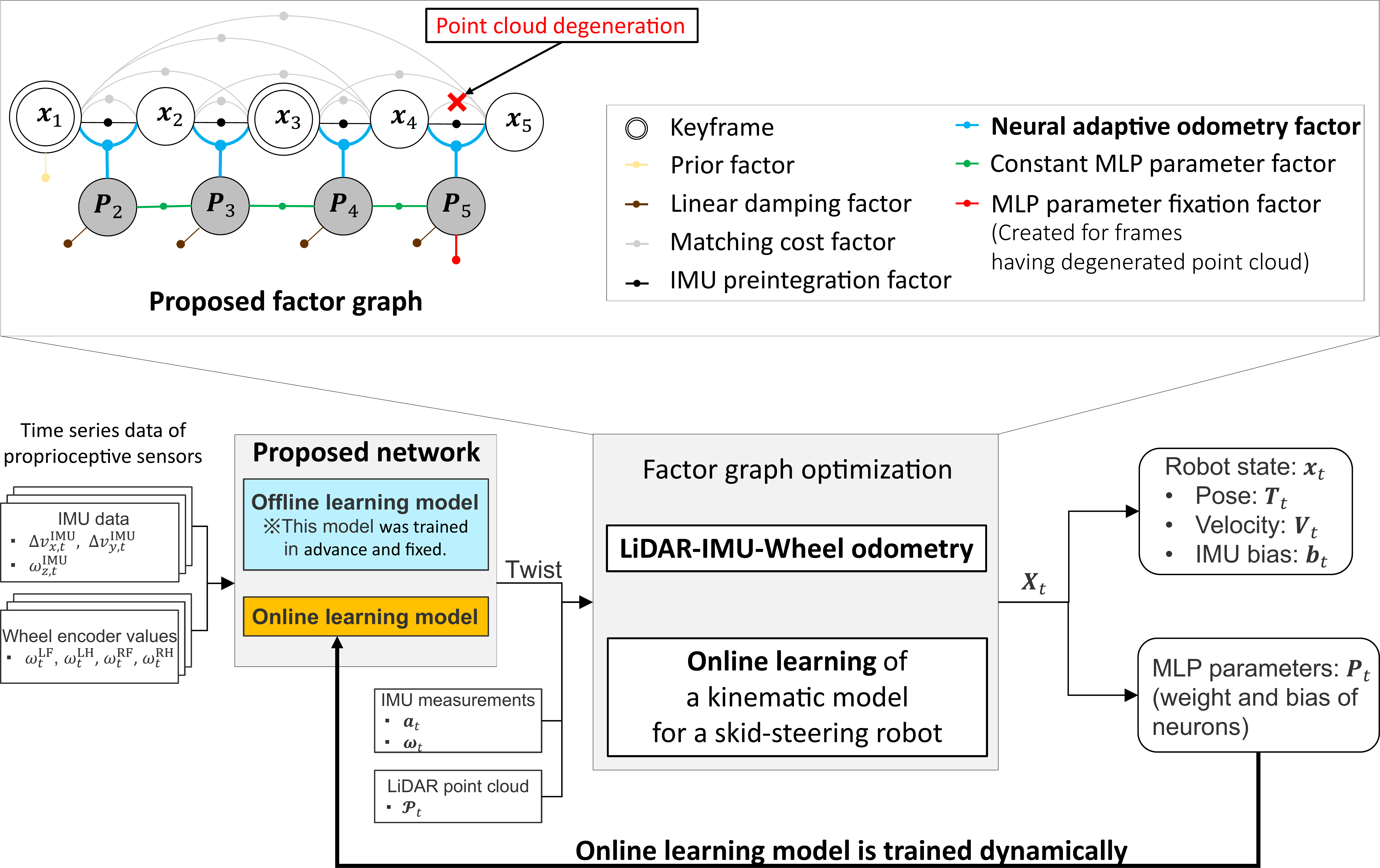}
  \caption{Overview of the proposed framework. We simultaneously solve online training of a neural network and LiDAR-IMU-wheel odometry on a unified factor graph based on a tightly coupled way. We designed the neural network to be divided into the \textit{offline learning model} and \textit{online learning model} to balance the computational cost and inference accuracy. We extended our previous work by introducing a \textit{neural adaptive odometry factor}, which provides motion constraints through a learning-based motion model and training of the \textit{online learning model}.
           }
  \label{fig:overview}
\end{figure*}
To estimate the time series of robot poses under severe point cloud degeneration and the nonlinear kinematic behavior of a skid-steering robot, we propose a tightly-coupled LiDAR-IMU-wheel odometry algorithm incorporating online training of the kinematic model, as seen in Figure~\ref{fig:overview}.
Odometry estimation and online training of the kinematic model are performed simultaneously on a unified factor graph to make all those constraints consistent. 

Online training of the neural network enables terrain-dependent features to adapt to the current terrain condition.
However, there is a trade-off between computational cost and inference accuracy for training the network; thus, training an accurate and large network online is infeasible.
Therefore, we divided the neural network into an \textit{offline learning model} and \textit{online learning model}.
The \textit{offline learning model} is trained in advance for expressing static terms (e.g., terrain-independent terms) of the kinematic models.
In contrast, the \textit{online learning model} is trained along with LiDAR-IMU-wheel odometry for expressing dynamically changing terms (e.g., terrain-dependent terms) of the kinematic models.

We define the state ${\bm X}_t$ to be estimated as 
\begin{align}
  \label{eq:state_declation}
  {\bm X}_t = [{\bm x}_t, {\bm P}_t], 
\end{align}
\begin{align}
  {\bm x}_t = [{\bm T}_t, {\bm v}_t, {\bm b}_t], 
\end{align}
where ${\bm x}_t$ is the robot state at time $t$; ${\bm T}_t~=~[{\bm R}_t | {\bm t}_t]~\in~SE(3)$ and ${\bm v}_t~\in~\mathbb{R}^3$ are the robot pose and velocity, respectively. ${\bm b}_t~=~[{\bm b}_t^a, {\bm b}_t^{\omega}]~\in~\mathbb{R}^6$ is the bias of the IMU acceleration ${\bm a}_t$ and the angular velocity ${\bm \omega}_t$.
${\bm P}_t$ indicates the weight and bias of each neuron to be learned online. 

We assume that ${\bm P}_t$ is trained online along with the robot states~${\bm x}_t$ in feature-rich environments before entering environments where LiDAR point clouds can degenerate. 
We also assume that the terrain condition does not drastically change while transitioning from feature-rich environments to another environment where LiDAR point clouds can degenerate. 
Once the online-trained neural network converges to a proper model that adapts to the current terrain condition, this model's inference becomes a reliable constraint for optimization and enables an accurate odometry estimation even in such severe environments.

Specifically, factor graph optimization is used for simultaneously performing LiDAR-IMU-wheel odometry and online training of the learning-based kinematic model, as shown in Figure~\ref{fig:overview}.
Details regarding each factor are explained in the following subsections.

\subsection{Matching cost factor}\label{subsec:MCF}
The matching cost factor~\cite{koide2021globally} references robot poses $\bm{T}_i$ and $\bm{T}_j$ to align their point cloud $\mathcal{P}_i$ and $\mathcal{P}_j$ based on voxelized GICP (VGICP) registration with GPU acceleration~\cite{koide2021voxelized}.
The VGICP algorithm treats point ${\bm p}_k$ in $\mathcal{P}_i$ as a Gaussian distribution; ${\bm p}_k = ({\bm \mu}_k, {\bm C}_k)$, where ${\bm \mu}_k$ and ${\bm C}_k$ are the mean and covariance matrix estimated from the neighbors of ${\bm p}_k$, respectively.
The target point cloud $\mathcal{P}_j$ is discretized by voxelization, and each voxel contains the average of the means and covariance matrices within its voxel points. 
This voxelization process enables a quick nearest-neighbor search through spatial hashing~\cite{teschner2003optimized}.
The matching cost $e^{\text {M}}({\bm T}_i, {\bm T}_j)$ is defined as follows:
\begin{align}
  \label{eq:vgicp}
  e^{\rm {M}}({\bm T}_i, {\bm T}_j) &= \sum_{p_k \in \mathcal{P}_i} e^{\text{\rm {D2D}}}({\bm p}_k, {\bm T}_i^{-1} {\bm T}_j), \\
  e^{\text{\rm {D2D}}} ({\bm p}_k, {\bm T}_{ij}) &= {\bm d}_k^\top ({\bm C}'_k + {\bm T}_{ij}{\bm C}_k{\bm T}_{ij}^\top)^{-1} {\bm d}_k,
\end{align}
where $e^{\text{\rm {D2D}}}$ is the distribution-to-distribution Mahalanobis distance between the input point ${\bm p}_k$ and the corresponding voxel ${\bm p}_k'~=~({\bm \mu}_k', {\bm C}_k')$, and ${\bm d}_k~=~{\bm \mu}_k' - {\bm T}_{ij} {\bm \mu}_k$ is the residual between ${\bm \mu}_k$ and ${\bm \mu}'_k$. 

As the implementation details, we conduct a deskewing process of input point cloud $\mathcal{P}_i$ by using IMU measurements for preprocessing. 
In addition, we manage the keyframes based on an overlap ratio between the input point cloud and the target point cloud.
If the overlap ratio is smaller than a threshold (e.g.,~$90\%$), we set the frame of the input point cloud as a keyframe.
As seen in the proposed factor graph in Figure~\ref{fig:overview}, we constrain each frame by matching cost factors with its last $\it N$ (e.g., 3) frames and previous keyframes to reduce the estimation drift.
More details of this implementation are given in~\cite{koide2022globally}. 

\subsection{IMU preintegration factor}\label{subsec:IPF}
The IMU preintegration factor is a constraint for the relative pose and the linear velocity between two consecutive frames by integrating IMU measurements (${\bm a}_i$ and ${\bm \omega}_i$).
This factor enables efficient optimization based on the preintegration technique, which avoids the recomputation of IMU measurement integration in every optimization iteration.
The IMU data are used to predict the sensor state at a time interval $\Delta t$ (i.e., the frequency of IMU measurements) as follows:
\begin{align}
  \label{eq:imu_evol_R}
  \hspace{-2.8mm}
  {\bm R}_{t + \Delta t} &= {\bm R}_t \exp \left( \left( {\bm \omega}_t - {\bm b}_t^{\omega} - {\bm \eta}_k^{\omega} \right) \Delta t \right), \\
  \label{eq:imu_evol_v}
  \hspace{-2.8mm}
  {\bm v}_{t + \Delta t} &= {\bm v}_t + {\bm g} \Delta t + {\bm R}_t \left( {\bm a}_t - {\bm b}_t^a - {\bm \eta}_t^a \right) \Delta t, \\
  \label{eq:imu_evol_p}
  \hspace{-2.8mm}
  {\bm t}_{t + \Delta t} &= {\bm t}_t + {\bm v}_t \Delta t + \frac{1}{2} {\bm g} \Delta t^2 + \frac{1}{2} {\bm R}_t \left( {\bm a}_t - {\bm b}_t^a - {\bm \eta}_t^a \right) \Delta t^2,
\end{align}
where $\bm g$ denotes the gravity vector and $\bm{\eta}_t^a$ and $\bm{\eta}_t^{\omega}$ are the white noise of the IMU measurement. 
We can calculate the relative robot motion $\Delta \bm{R}_{ij}$, $\Delta \bm{v}_{ij}$, and $\Delta \bm{t}_{ij}$ by integrating $\bm{R}_{t+\Delta t}$, $\bm{v}_{t+\Delta t}$, and $\bm{t}_{t+\Delta t}$, respectively, between time steps $i$ and $j$. 
The error $e^{\rm {IMU}}(\bm{x}_i, \bm{x}_j)$ between consecutive frame states ${\bm x}_i$ and ${\bm x}_j$ is finally defined as follows:
\begin{equation}
  \begin{aligned}
    \hspace{-2.8mm}
    e^{\text{IMU}}\left(\bm{x}_i, \bm{x}_j\right) & =\left\|\log \left(\Delta \bm{R}_{ij}^\top \bm{R}_i^\top \bm{R}_j\right)\right\|^2 \\
    & +\left\|\Delta \bm{t}_{ij}-\bm{R}_i^\top\left(\bm{t}_j-\bm{t}_i-\bm{v} \Delta t_{ij}-\frac{1}{2} \bm{g} \Delta t_{ij}^2\right)\right\|^2 \\
    & +\left\|\Delta \bm{v}_{ij}-\bm{R}_i^\top\left(\bm{v}_j-\bm{v}_i-\bm{g} \Delta t_{ij}\right)\right\|^2 .
  \end{aligned}
  \label{eq:imu_error}
  \end{equation}
  
More details are given in~\cite{forster2016manifold}. 
The IMU preintegration factor enables robust estimation for rapid motion and short-time degeneracy of point clouds. 
Furthermore, this factor can reduce the estimation drift of poses from six to four DoFs~\cite{qin2018vins}.

\subsection{Neural adaptive odometry factor}\label{FoooNET}
We propose a neural network that infers the 2D twist of a skid-steering robot with proprioceptive sensors, namely encoder and IMU. 
This neural network consists of two sub-networks: 1) \textit{offline learning model} describing static features (e.g., terrain-independent parameters) of the kinematic model and 2) \textit{online learning model} expressing dynamically changing features (e.g., terrain-dependent terms).
The \textit{offline learning model} is obtained by batch learning before odometry estimation is conducted.
In contrast, the \textit{online learning model} is trained online by factor graph optimization along with the robot state ${\bm x}$.
To conduct the online learning by factor graph optimization, we propose the neural adaptive odometry factor to constrain the robot pose ${\bm T}$, the linear velocity ${\bm v}$, and the parameter vector ${\bm P}$ concatenating the weight and bias of each neuron to be trained online.

\begin{figure}[tb]
  \centering
  \includegraphics[width=0.65\linewidth]{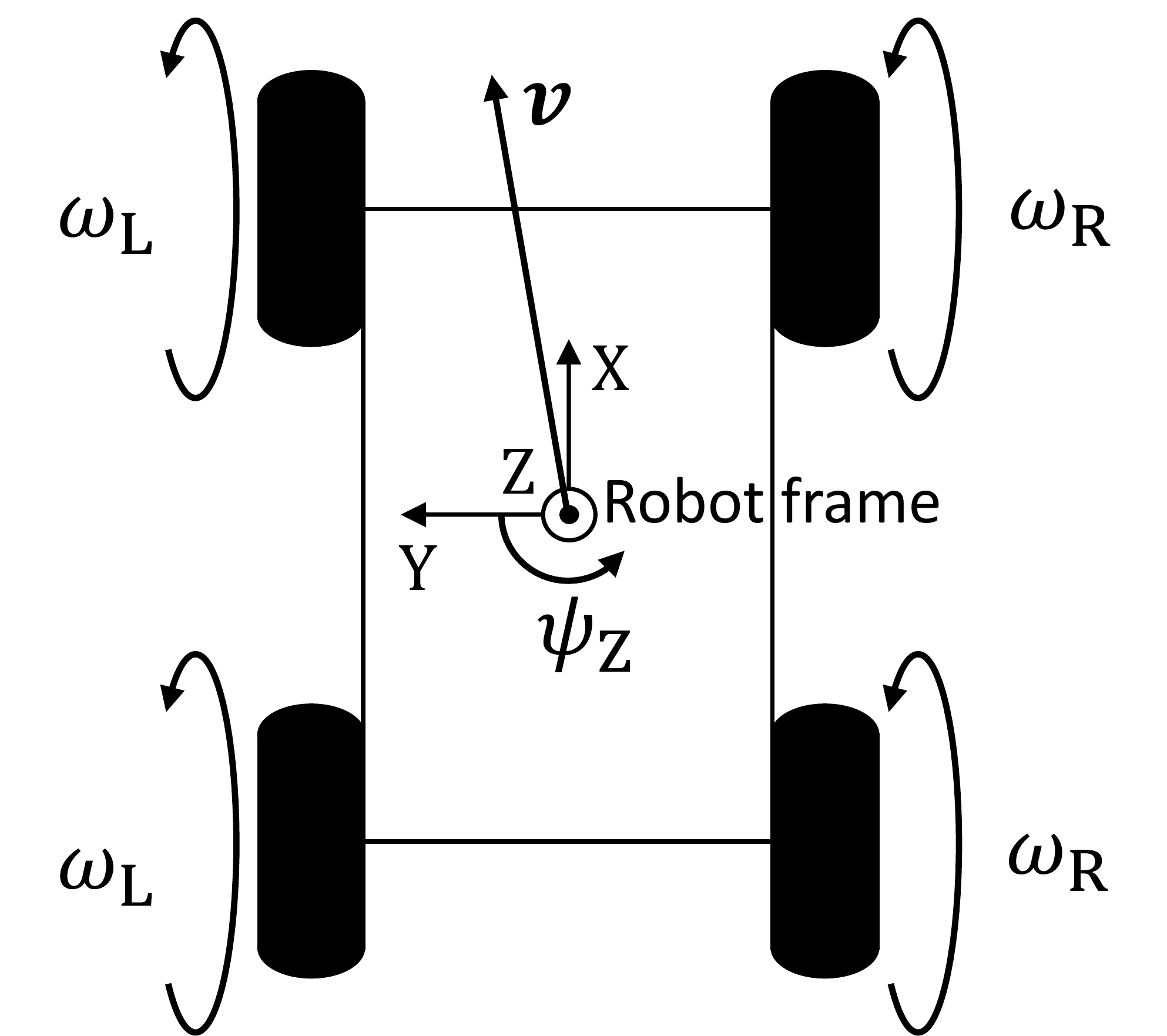}
  \caption{Kinematic model of a skid-steering robot.}
  \label{fig:wheel_robot_description}
\end{figure}
\subsubsection{Fundamental kinematic model for wheeled robots}
First, we introduce the fundamental formulation of a kinematic model for wheeled robots.
In the case of an ideal condition (e.g., wheel slippage and lateral motion do not occur, and wheels and terrain are not deformed), the robot twist motion is described by a simple kinematic equation:
\begin{align} 
  \begin{bmatrix}
    v_{x} \\
    v_{y} \\
    \psi_{z} \\
  \end{bmatrix}
    = 
    J \bm \omega
    =
    \begin{bmatrix}
      R/2 & R/2 \\
      0 & 0 \\
      -R/B & R/B \\
    \end{bmatrix}
    \begin{bmatrix}
      \omega_{\rm L} \\
      \omega_{\rm R} \\
  \end{bmatrix},
  \label{eq:ideal_wo}
\end{align}
where ${\bm v}=[v_{x} ~v_{y}]^\top$ is the 2D translational velocity; ${\psi_{z}}$ is the angular velocity around the z-axis of the robot frame; 
$R$ and $B$ indicate the wheel radius and wheelbase, respectively;
\mbox{${\bm \omega} = [\omega_{\rm L} ~\omega_{\rm R}]^\top$} is the angular velocity of the left and right wheels, as shown in Figure~\ref{fig:wheel_robot_description}.
${\bm J}$ is a \(3 \times 2\) matrix that describes the relationship between $[v_{x}~v_{y}~{\psi_{z}}]^\top$ and $\bm \omega$.
${\bm J}$ is constant in the case of ideal conditions.
As seen in Eq.~\ref{eq:ideal_wo}, the relationship between the robot twist $[v_{x}~v_{y}~{\psi_{z}}]$ and the wheels' angular velocity ${\bm \omega}$ is clearly linear. 

In the case of skid-steering robots, Eq.~\ref{eq:ideal_wo} does not fully capture the robot motion, because the skid-steering robot rotates by skidding its wheels depending on the difference between the angular velocities of the left and right wheels.
Thus, wheel slippage and lateral motion $v_{y}$ must be considered for skid-steering robots.
However, incorporating these terms makes the robot twist motion model more complex because this rigorous model depends on terrain-dependent parameters (e.g., friction coefficients).
Many works modeled the ${\bm J}$ matrix for expressing the complex motion for skid-steering robots\cite{anousaki2004dead,le1997estimation,mandow2007experimental,wang2015analysis,rabiee2019friction}; however, these linear models cannot capture nonlinear kinematic behavior (e.g., large wheel slippage such as drifting).

\subsubsection{Proposed neural network structure}\label{FoooNET_structure}
\begin{figure*}[tb]
  \centering
  \includegraphics[width=1\linewidth]{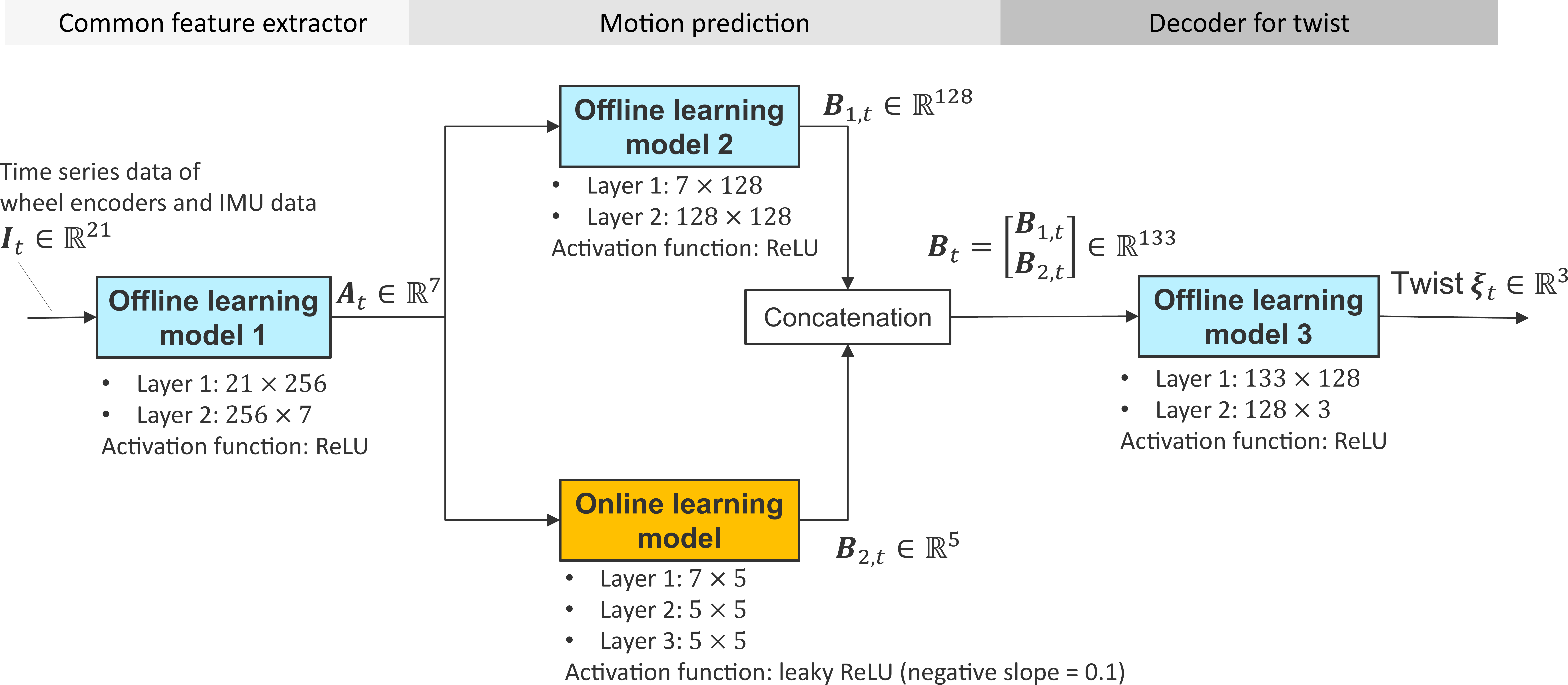}
  \caption{Proposed neural network structure. The network outputs twist from the time series data of wheel encoders and IMU data, through the following three layers: 1) \textit{Common feature extractor}, 2) \textit{Motion prediction}, and 3) \textit{Decoder for twist}.
  We divided the second layer into two blocks to be trained online and offline for capturing terrain-dependent and -independent properties of the kinematic model, respectively.
  The number of neurons in the \textit{online learning model} was set to 100 ($(7\times 5 + 5) + (5\times 5 + 5) + (5\times 5 + 5)$); thus, the dimension of MLP parameter $P$ is 100 in our implementation.
           }
  \label{fig:FoooNET}
\end{figure*}
We implicitly express the complex motion model with nonlinearity based on a nonlinear neural network.
The proposed neural network must be trained online to adapt to each terrain condition because the kinematic model is subject to terrain-dependent parameters.
As shown in Figure~\ref{fig:FoooNET}, we designed the network with three layers: 1) the first layer extracts common features for motion prediction from the time series data of wheel encoders and IMU data, 2) the second layer predicts motion in the latent space, and 3) the third layer decodes the features into the twist.
We divided the second motion prediction layer into two blocks to be trained online and offline for capturing terrain-dependent and -independent properties of the kinematic model, respectively.
We designed the proposed neural network such that the \textit{online learning model} represents terrain-dependent properties with fewer neurons compared to those in the \textit{offline learning models} for balancing inference accuracy and computational costs.
In our implementation, the number of neurons in the \textit{online learning model} was set to 100; that is, the dimension of MLP parameters vector ${\bm P}_t$ is 100.

The input data of the proposed neural network is the time series data measured by the wheel encoders and IMU, namely, all-wheel rotational velocities {$\omega _{t} ^{\rm{LF}}$, $\omega _{t} ^{\rm{LH}}$, $\omega _{t} ^{\rm{RH}}$, $\omega _{t} ^{\rm{RF}}$}, incremental velocity $\Delta v _{x, t}^{ \rm IMU}$, $\Delta v _{y, t}^{ \rm IMU}$, and an angular velocity $\omega _{z, t}^{ \rm IMU}$.
Note that $\Delta v _{x, t}^{ \rm IMU}$, $\Delta v _{y, t}^{ \rm IMU}$, and $\omega _{z, t}^{ \rm IMU}$ are described in the robot frame.
We define $\bm {i}_t$ as a concatenated vector of those seven types of sensor data as follows:
\begin{align}
  \label{eq:input_NN_i}
  \bm {i}_{t} &=[{\omega _{t} ^{\rm{LF}}, \omega _{t} ^{\rm{LH}}, \omega _{t} ^{\rm{RH}}, \omega _{t} ^{\rm{RF}}, \Delta v _{x, t}^{ \rm IMU}, \Delta v _{y, t}^{ \rm IMU}, \omega _{z, t}^{ \rm IMU}}].
\end{align}
We concatenate consecutive $N_{\rm w}$ (e.g., 3) frames of $\bm {i}_{t}$ to compose an input vector $I_t = \rho([ i_t, i_{t-1}, ... ])$, where $\rho$ is the standardization operation.
The proposed neural network outputs twist $\bm {\xi}_t \in ~\mathbb{R}^3$ through the input data $\bm {I}_{t}$ as shown in Figure~\ref{fig:FoooNET}.

We empirically chose the current number of layers and neurons from various network configurations.
For example, we faced overfitting particularly in the case of pivot turn inducing large wheel slippage when we increased the number of layers of offline learning model 1, 2, and 3 to three layers.
We also observed that the validation accuracy was not improved if the number of neurons was increased (e.g., 1024, 512). We consider this because our network task (2D twist inference through IMU and wheel encoder) is relatively simpler than other complex tasks (e.g., image classification) that require significantly large models due to the complexity of the objective.

\begin{figure*}[tb]
  \centering
  \includegraphics[width=1\linewidth]{figs/dataset_summary.pdf}
  \caption{Each environment of datasets used for offline learning: \#1,~indoor flat floor; \#2,~wood tiles; \#3,~bricks; \#4, stone tiles; \#5,~cement tiles; \#6,~grass; \#7, dry asphalt; and \#8,~wet asphalt. The robot was operated with various speeds (0 to 3~\si{m}/\si{s}) and trajectories (e.g., straight, curved, and pivot turns) in each dataset. Environments in all datasets were surrounded by many features, and thus the result of LiDAR-IMU odometry can be used as reference data for offline learning.}
  \label{fig:dataset}
\end{figure*}

\subsubsection{Offline learning procedure}
In the offline training phase, we aim to train \textit{offline learning model~1, 2, and 3} in Figure~\ref{fig:FoooNET} to capture the robot behavior while conditioning them to the \textit{online learning model}, which will be trained online to encode dynamic environment information.
We assume that the parameters of the \textit{online learning model} do not drastically change when the environment is the same.
Thus, we conduct offline learning such that, while \textit{offline learning model~1, 2, and 3} consist of the same parameters for all environments, the \textit{online learning model} consists of different parameters for each environment.

\begin{figure}[tb]
  \centering
  \includegraphics[width=1\linewidth]{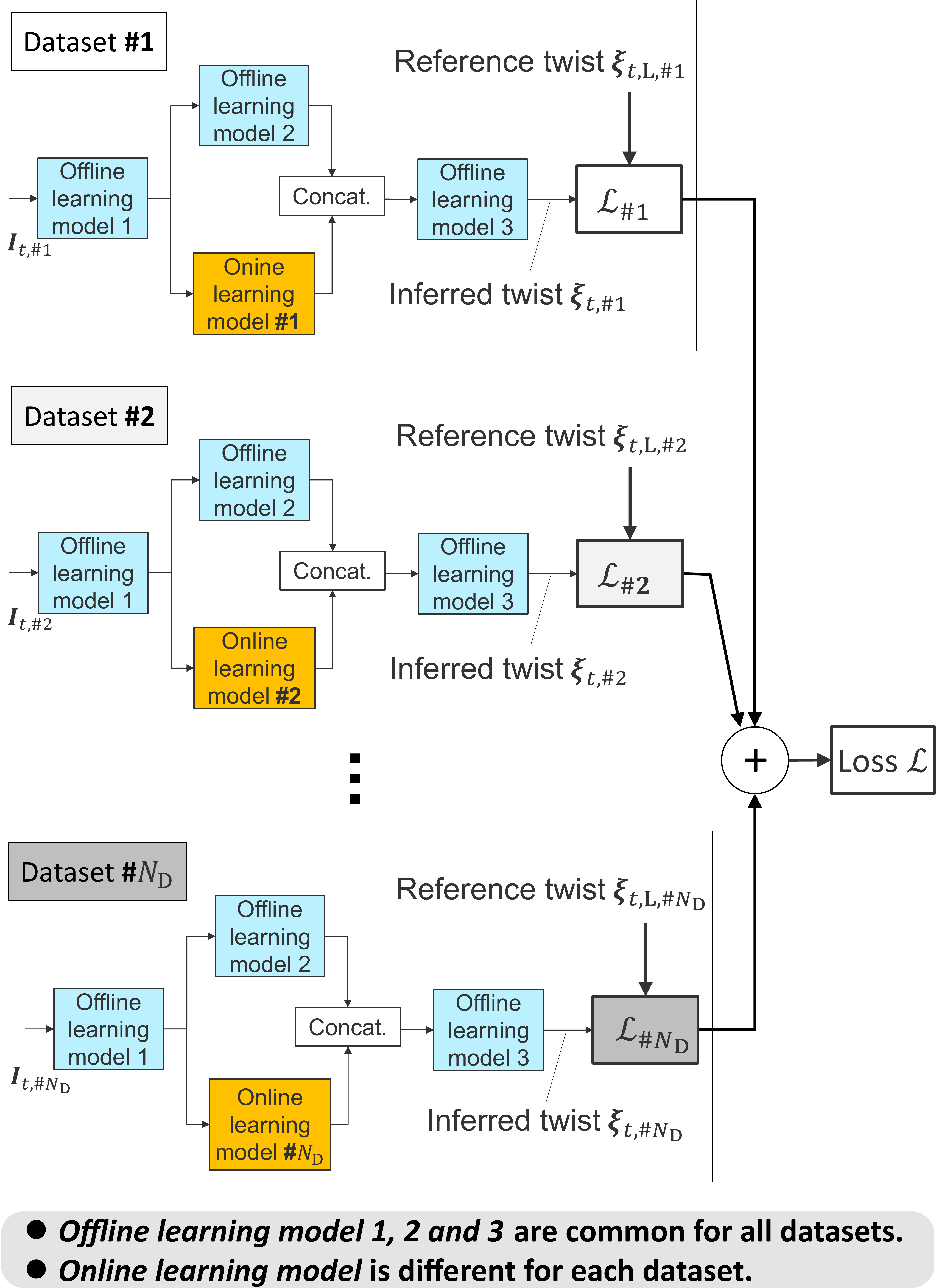}
  \caption{Offline learning procedure for obtaining \textit{offline learning model~1, 2, and 3}. The loss function $\mathcal{L}$ is defined as the sum of all loss functions from $\mathcal{L}_{\# 1}$ to $\mathcal{L}_{\#N_{\rm D}}$ obtained with all datasets. In our implementation, we set the number of datasets $N_{\rm D}$ to 8, as shown in Figure~\ref{fig:dataset}. 
           }
  \label{fig:offline_and_online}
\end{figure}
To conduct offline learning in this framework, we prepared a dataset including $N_{\rm D}$ (e.g., 8) sequences in different environments, such as those shown in Figure~\ref{fig:dataset}.
As seen in Figure~\ref{fig:offline_and_online}, each loss function $\mathcal{L}_{\# n}$ is defined for each dataset, and the offline learning is executed to minimize the sum of the loss functions $\mathcal{L}$.
We train the proposed network so that, while \textit{offline learning model~1, 2, and 3} are common for all sequences, the \textit{online learning model} is different for each sequence.
Let \mbox{$\bm{M}_{\text{off}}$} and \mbox{$\bm{M}_{\text{on},n}$} denote the training parameters of the offline learning model and online learning model in our offline learning phase, respectively.
Note that \mbox{$\bm{M}_{\text{on},n}$} is defined by the number of environment datasets (i.e., \mbox{$N_{\rm D}$}).  
To meet the aforementioned offline learning condition, we minimize a loss function \mbox{$\mathcal{L}$} defined in Eq.~\mbox{\ref{eq:all_loss}} such that while \mbox{$\bm{M}_{\text{off}}$} is common for all environments, \mbox{$\bm{M}_{\text{on},n}$} consists of different parameters for each environment.
\begin{equation}
  \label{eq:all_loss}
  \begin{aligned}
    &\mathcal{L}\left(\boldsymbol{M}_{\text {off}}, \boldsymbol{M}_{\text {on}, \# 1}, \ldots, \boldsymbol{M}_{\text {on}, \# N_{\mathrm{D}}}\right) = \sum_{n=1}^{N_{\mathrm{D}}} \mathcal{L}_{\# n}\left(\boldsymbol{M}_{\text {off}}, \boldsymbol{M}_{\text {on}, \# n}\right),
  \end{aligned}
\end{equation}
\vspace{-8mm}
\begin{equation}
  \label{eq:each_loss}
  \begin{aligned}
    &\mathcal{L}_{\# n}\left(\boldsymbol{M}_{\text {off}}, \boldsymbol{M}_{\text {on}, \# n}\right)=\\& \frac{1}{n_{\rm D}} \sum_{t=1}^{n_{\rm D}}\left(e_{t, \text { trans }}\left(\boldsymbol{M}_{\text {off}}, \boldsymbol{M}_{\text {on}, \# n}\right)+w e_{t, \text { rot }}\left(\boldsymbol{M}_{\text {off}}, \boldsymbol{M}_{\text {on}, \# n}\right)\right),
  \end{aligned}
\end{equation}
where \mbox{$\mathcal{L}_{\# n}\left(\boldsymbol{M}_{\text {off}}, \boldsymbol{M}_{\text {on}, \# n}\right)$} is the Mean Square Error (MSE) of the estimated twist for each dataset;
\mbox{$n_{\rm D}$} is the number of reference data in each dataset;
\mbox{$e_{t, \text { trans }}\left(\boldsymbol{M}_{\text {off}}, \boldsymbol{M}_{\text {on}, \# n}\right)$} and \mbox{$e_{t, \text { rot }}\left(\boldsymbol{M}_{\text {off}}, \boldsymbol{M}_{\text {on}, \# n}\right)$} are the square error of the translation and rotation components of the twist \mbox{$\bm {\xi}_t$}, respectively;
and \mbox{$w$} is a weight (e.g., 32) for adjusting the scale of translation and rotation.

We used the Adam optimizer~\cite{KingBa15} with a learning rate of $10^{-4}$ and 3000 epochs.
The batch size was set to 128 in our implementation.
The number of layers and activation functions for each MLP are shown in Figure~\ref{fig:FoooNET}.
We utilized LiDAR-IMU odometry with an omnidirectional field-of-view (FOV) LiDAR (Livox MID-360) to obtain the reference data of the twist $\bm {\xi}_{t, \rm{L}}$.
LiDAR-IMU odometry is highly accurate in feature-rich environments, such as those in the datasets we used (Figure~\ref{fig:dataset}).
The robot also moved locally in each environment to always remain on the local map, and thus the LiDAR-IMU odometry estimation did not accumulate errors.
In each dataset, we manually operated the robot at a range of speeds (0 to 3~\si{m}/\si{s}) and various motions (straight and curved trajectories, and pivot turns).
The offline training takes 2.5 hours on a desktop PC equipped with the NVIDIA GeForce RTX 4080 graphic card and intel core i7-13700KF processor.

\subsubsection{Error derivation of neural adaptive odometry factor for online learning}\label{FoooNETOnline}
Unlike the \textit{offline learning model}, the \textit{online learning model} is adaptively updated to better capture the dynamic environment features obtained by the \textit{neural adaptive odometry factor} through factor graph optimization.
This factor provides not only motion constraints but also training for the \textit{online learning model}, and thus the parameter vector $\bm P$ concatenating weights and biases of neurons is trained online via factor graph optimization on a unified factor graph along with robot state $\bm x$.

First, we describe the pose error $e^{\rm {NN}}({{\bm T}_{i-1}, {\bm T}_i, {\bm P}_i})$. 
We extend the output twist $\bm {\xi}_i \in ~\mathbb{R}^3$ into $\bm {\xi}_{i,\rm {6DoF}} \in \mathbb{R}^6$ by substituting 0 into the translational z and rotational x, y of $\bm {\xi}_{i,\rm {6DoF}}$. 
We calculate the displacement by integrating $\bm {\xi}_{i,\rm {6DoF}}$ in the time step $\Delta t_{i}$.
The pose error $e^{\rm {NN}}({{\bm T}_{i-1}, {\bm T}_i, {\bm P}_i})$ is defined as follows:
\begin{align}
  \label{eq:pose_cost}
  e^{\rm {NN}}({{\bm T}_{i-1}, {\bm T}_i, {\bm P}_i}) &= ({{\bm r}_{i}^{\rm {NN}_{p}}})^\top {{\bm C}_{i}^{\rm {NN}}}^{-1} {\bm r}_{i}^{\rm {NN}_{p}}, \\
  \bm r_{i}^{\rm {NN}_{p}} &= \log ( {{\bm T}_{i-1}^{-1} {\bm T}_i \exp (\bm {\xi}_{i,\rm {6DoF}} \Delta t_{i}) ^{-1}} ),
\end{align}
where ${{\bm C}_{i}^{\rm {NN}}}^{-1}$ is the covariance matrix of the displacement.
We estimate ${{\bm C}_{i}^{\rm {NN}}}^{-1}$ online with the Kalman filter (as described in our previous work~\cite{okawara2024tightly}) to obtain the uncertainty of the 3D motion.
This method can adaptively set the covariance matrix based on terrain roughness.

Next, we define the linear velocity error $e^{\rm {NN}}(\bm {T}_{i-1}, \bm {v}_{i-1}, {\bm P}_i)$.
We extend the twist $\bm {\xi}_i \in \mathbb{R}^3$ into the 3D linear velocity $\bm {v}_{i}^{\rm {NN}} \in \mathbb{R}^3$ by substituting 0 into the z element of $\bm {v}_{i}^{\rm {NN}}$.
The linear velocity error $e^{\rm {NN}}(\bm {T}_{i-1}, \bm {v}_{i-1}, {\bm P}_i)$ is defined as follows:
\begin{align}
  \label{eq:linear_vel_cost}
  e^{\rm {NN}}(\bm {T}_{i-1}, \bm {v}_{i-1}, {\bm P}_i) &= ({{\bm r}_{i}^{\rm {NN_V}}})^\top {{\bm C}_{i}^{\rm {NN_V}}}^{-1} {\bm r}_{i}^{\rm {NN_V}}, \\
  \bm r_{i}^{\rm {NN_V}} &= \bm {v}_{i-1} - {\bm R}_{i-1}  \bm {v}_{i}^{\rm {NN}},
\end{align}
where ${{\bm C}_{i}^{\rm {NN_V}}}^{-1}$ is the covariance matrix of the linear velocity.
We set all diagonal elements of ${{\bm C}_{i}^{\rm {NN_V}}}^{-1}$ to $10^{-6}$.

Finally, the error of the neural adaptive odometry factor $e^{\rm {NN}}({\bm T}_{i-1}, {\bm T}_i, \bm {v}_{i-1}, {\bm P}_i)$ is defined as the sum of $e^{\rm {NN}}({{\bm T}_{i-1}, {\bm T}_i}, {\bm P}_i)$ and $e^{\rm {NN}}(\bm {T}_{i-1}, \bm {v}_{i-1}, {\bm P}_i)$ as follows:
\begin{align}
  \label{eq:FoooNET_cost}
  e^{\rm {NN}}({\bm T}_{i-1}, {\bm T}_i, \bm {v}_{i-1}, {\bm P}_i) &= e^{\rm {NN}}({{\bm T}_{i-1}, {\bm T}_i}, {\bm P}_i) \notag \\
  & + e^{\rm {NN}}(\bm {T}_{i-1}, \bm {v}_{i-1}, {\bm P}_i).
\end{align}

\subsection{Constant MLP parameter factor}\label{Constant}
\label{subsec:constant_factor}
We assume that dynamic environment parameters do not change drastically over short time periods and create a \textit{constant MLP parameter factor} between consecutive states with zero-mean Gaussian noise.
The \textit{constant MLP parameter factor} constrains the MLP parameters of consecutive time steps to become mostly identical under the random walk assumption.
The error~$e^{\rm {C}}({\bm P}_{i-1}, {\bm P}_i)$ of this factor is defined as follows:
\begin{align}
  \label{eq:const_MLP_error}
  e^{\rm {C}}({\bm P}_{i-1}, {\bm P}_i) &= {\bm {r}_{i}^{\rm {C}}}^\top {{\bm C}_{i}^{\rm {C}}}^{-1} {\bm {r}_{i}^{\rm {C}}},\\
  \bm {r}_{i}^{\rm {C}} &= {\bm P}_i - {\bm P}_{i-1}.
\end{align}
In our implementation, we set all diagonal elements of the covariance matrix~${{\bm C}_{i}^{\rm {c}}}^{-1}$ of this factor to $10^{-6}$.

\subsection{MLP parameter fixation factor}\label{MLPConst}
The neural adaptive odometry factor facilitates the online learning for ${\bm P}_i$ and serves as a motion constraint.
Therefore, online learning can be unstable when point cloud-based constraints become unreliable due to point cloud degeneration.
To resolve this problem, we introduce a \textit{MLP parameter fixation factor} for ${\bm P}_i$ whenever point cloud degeneration is detected in the current frame.

After the factor graph is optimized, we detect point cloud degeneration by analyzing the linearized system of the matching cost factor (point cloud-based constraint).
This procedure involves two steps: 1) calculating the Hessian matrix of a matching cost factor between the latest and last frames and 2) determining degeneration if the minimum eigenvalue of the Hessian matrix is smaller than a certain threshold (i.e., the Hessian matrix is not positive definite). 
If the current point cloud $\mathcal{P}_i$ is determined as degenerate, we create the \textit{MLP parameter fixation factor} to ensure that ${\bm P}_i$ is retained to the values ($\tilde{\bm P}_{i-1}$) just before degeneration occurs to avoid numerically unstable optimization.
The error~$e^{\rm {F}}({\bm P}_i)$ of this factor is defined as follows:
\begin{align}
  \label{eq:MLPConstEq}
  e^{\rm {F}}({\bm P}_i)&= \begin{cases}{\bm {r}_{i}^{\rm {F}}}^\top {{\bm C}_{i}^{\rm {F}}}^{-1} {\bm {r}_{i}^{\rm {F}}}, &  { \text{if}~ \mathcal{P}_i ~ \text{is degenerated} } \\ 0, & \text { otherwise }\end{cases}\\
  \bm {r}_{i}^{\rm {F}} &= {\bm P}_i - \tilde{\bm P}_{i-1}.
\end{align}
Note that $e^{\rm {F}}({\bm P}_i)$ become $0$ when the current point cloud $\mathcal{P}_i$ is not detected as degenerated.
$\tilde{\bm P}_{i-1}$ is a fixed MLP parameter at time $i-1$, and not optimized.
All diagonal elements of the covariance matrix ${{\bm C}_{i}^{\rm {F}}}^{-1}$ of this factor are set to $10^{-12}$ to represent hard constraints in our implementation. 
We also use similar factors for the IMU bias to ensure stable optimization.

\subsection{Implementation details}\label{Implementation}
In summary, the objective function $e^{\text {ALL}}$ of our system is defined as follows:
\begin{align}
  \label{eq:FGO}
  e^{\text{ALL}} &= \underbrace{e^{\text{Prior}}(\bm{x}_1)}_{\text{Prior factor}} + 
  \sum_{j\in{\mathcal{K},~i=2 }} \underbrace{e^{\text{M}}({\bm T}_j, {\bm T}_i)}_{\text{Matching cost factor}} \notag \\
                 & + \sum_{i=2} \underbrace{e^{\text{IMU}}(\bm{x}_{i-1}, \bm{x}_i)}_{\text{IMU preintegration factor}} \notag \\
                 & + \sum_{i=2} \underbrace{e^{\text{NN}}({\bm T}_{i-1}, {\bm T}_i, \bm {v}_{i-1}, {\bm P}_i)}_{\text{Neural adaptive odometry factor}} \notag \\
                 & + \sum_{i=3} \underbrace{e^{\text{C}}({\bm P}_{i-1}, {\bm P}_i)}_{\text{Constant MLP parameter factor}} \notag \\
                 & + \sum_{i=3} \underbrace{e^{\text{F}}({\bm P}_i)}_{\text{MLP parameter fixation factor}}.
\end{align}
As seen in the factor graph in Figure~\ref{fig:overview}, the error $e^{\text{Prior}}(\bm{x}_1)$ of the prior factor constrains the initial robot state ${\bm x}_1$ to the fixed reference frame. 
Similar to the Levenberg-Marquardt algorithm, we add a constant damping element to the Hessian matrix related to ${\bm P}$ at each timestep to help the optimization become more stable because ${\bm P}$ is a large variable with 100 dimensions, which can cause numerically unstable optimization.
As stated in Section \ref{subsec:MCF}, $\mathcal{K}$ is a set of point clouds as follows: 1) the last three point clouds ${\mathcal{P}_{i-1}, \mathcal{P}_{i-2}, \mathcal{P}_{i-3}}$ and 2) keyframes.

We implemented our robot system using ROS~2 Humble on Ubuntu~22.04 for sensor data collection and testbed control.
We used \href{https://github.com/borglab/gtsam}{GTSAM} to implement the factor graph and the iSAM2~\cite{kaess2012isam2} optimizer to incrementally optimize the factor graph (i.e., $e^{\text {ALL}}$).
We also used \href{https://github.com/pytorch/pytorch}{PyTorch} and \href{https://pytorch.org/get-started/locally/}{LibTorch} for implementing offline learning procedures~(Figure~\ref{fig:offline_and_online}) and the neural adaptive odometry factor, respectively.

\section{Experimental results}\label{results}
First, we verified that the network we developed can properly adapt to each terrain condition (Section~\ref{subsection:FoooNETresults}).
Second, we demonstrated that the proposed odometry estimation algorithm can deal with severe point cloud degeneration and large wheel slippage (i.e., nonlinearity in the kinematic model of the skid-steering robot) (Section~\ref{subsection:LIWOresults}).

\subsection{Verification evaluation of the proposed neural network}\label{subsection:FoooNETresults}
To demonstrate that the proposed network (Figure~\ref{fig:FoooNET}) and training approach (Figure~\ref{fig:offline_and_online}) enable robust odometry estimation in different terrains, we conducted an ablation study that evaluated the wheel odometry estimation of the network only, that is, without LiDAR and IMU constraints.
We compared the proposed trained network (\textit{Ours}) with other networks trained through two offline learning procedures as follows:
\begin{enumerate}
  \item \textit{Identical network}: In this model, a single \textit{online learning model} (i.e., simple batch learning-based model) was fitted to all the datasets; thus, features depending on terrain types could not be expressed accurately.
  \item \textit{Improper dataset pair}: This setting used the parameters of the \textit{online learning model} that was trained on a sequence different from that of the testing environment. We used the following two combinations of the \textit{online learning model}:
  \begin{enumerate}
    \item Case 1: Training=indoor flat floor, Testing=wood tiles
    \item Case 2: Training=indoor flat floor, Testing=grass
  \end{enumerate} 
  As seen in Figure~\ref{fig:dataset}, in Case 1 (indoor flat floor (\#1) and wood tiles (\#2)), the terrain differences were slight, whereas in Case 2 (indoor flat floor (\#1) and grass (\#6)), the differences in terrain conditions were significant. The indoor flat floor and wood tiles were flat and hard, whereas the grass was uneven and soft.
\end{enumerate} 
\begin{table*}[t]
  \centering
  \caption{RTEs per \SI{1}{m} for the ablation study on the \textit{Identical network} case on eight terrains. While the proposed method (Ours) trained the proposed neural network to adapt to the current environment (i.e., the \textit{online learning model} was trained for each different dataset), the \textit{Identical network} case trained its model to fit to all the datasets.}
  \label{tab:ablation_study_of_identicalMLP3}
  \begin{threeparttable}
    \setlength{\tabcolsep}{0.75mm} 
    \begin{tabular}{cc|cc|cc|cc|cc|cc|cc|cc|cc|cc}
      \hline
      \multicolumn{2}{c|}{} & 
      \multicolumn{2}{c|}{\#1} & 
      \multicolumn{2}{c|}{\#2} & 
      \multicolumn{2}{c|}{\#3} & 
      \multicolumn{2}{c|}{\#4} & 
      \multicolumn{2}{c|}{\#5} & 
      \multicolumn{2}{c|}{\#6} & 
      \multicolumn{2}{c|}{\#7} & 
      \multicolumn{2}{c|}{\#8} &
      \multicolumn{2}{c}{Mean}\\ \cline{3-20}

      \multicolumn{2}{c|}{\multirow{-2}{*}{Method/Seq.}} & 
      $t_{\rm{RTE}}$\tnote{1} & \cellcolor[HTML]{EFEFEF}$r_{\rm{RTE}}$ & 
      $t_{\rm{RTE}}$ & \cellcolor[HTML]{EFEFEF}$r_{\rm{RTE}}$ & 
      $t_{\rm{RTE}}$ & \cellcolor[HTML]{EFEFEF}$r_{\rm{RTE}}$ & 
      $t_{\rm{RTE}}$ & \cellcolor[HTML]{EFEFEF}$r_{\rm{RTE}}$ & 
      $t_{\rm{RTE}}$ & \cellcolor[HTML]{EFEFEF}$r_{\rm{RTE}}$ & 
      $t_{\rm{RTE}}$ & \cellcolor[HTML]{EFEFEF}$r_{\rm{RTE}}$ & 
      $t_{\rm{RTE}}$ & \cellcolor[HTML]{EFEFEF}$r_{\rm{RTE}}$ & 
      $t_{\rm{RTE}}$ & \cellcolor[HTML]{EFEFEF}$r_{\rm{RTE}}$ & 
      $t_{\rm{RTE}}$ & \cellcolor[HTML]{EFEFEF}$r_{\rm{RTE}}$ \\ \hline \hline
      
      \multicolumn{2}{c|}{Ours} & 
      \textbf{0.017} & \cellcolor[HTML]{EFEFEF}{0.24} & 
      \textbf{0.016} & \cellcolor[HTML]{EFEFEF}{0.25} & 
      \textbf{0.016} & \cellcolor[HTML]{EFEFEF}{0.21} & 
      \textbf{0.016} & \cellcolor[HTML]{EFEFEF}{0.20} & 
      \textbf{0.019} & \cellcolor[HTML]{EFEFEF}{0.28} & 
      \textbf{0.038} & \cellcolor[HTML]{EFEFEF}{0.31} & 
      \textbf{0.021} & \cellcolor[HTML]{EFEFEF}{0.24} &
      \textbf{0.025} & \cellcolor[HTML]{EFEFEF}{0.23} &
      \textbf{0.021} & \cellcolor[HTML]{EFEFEF}{0.25}\\ 
      
      \multicolumn{2}{c|}{Identical network} & 
      0.025 & \cellcolor[HTML]{EFEFEF}0.22 & 
      0.023 & \cellcolor[HTML]{EFEFEF}0.20 & 
      0.034 & \cellcolor[HTML]{EFEFEF}0.20 & 
      0.018 & \cellcolor[HTML]{EFEFEF}0.17 & 
      0.027 & \cellcolor[HTML]{EFEFEF}0.36 & 
      0.051 & \cellcolor[HTML]{EFEFEF}0.27 & 
      0.030 & \cellcolor[HTML]{EFEFEF}0.21 &
      0.038 & \cellcolor[HTML]{EFEFEF}0.18 &
      0.031 & \cellcolor[HTML]{EFEFEF}0.23\\ \hline

    \end{tabular}
    \begin{tablenotes}[flushleft]
      \item[1] $t_{\rm{RTE}}$ and $r_{\rm{RTE}}$ are the translation [\si{m}] and rotation [\si{deg}/\si{m}] RTEs, respectively.
    \end{tablenotes}
  \end{threeparttable}
\end{table*}
\begin{table}[t]
  \centering
  \caption{RTEs per \SI{1}{m} for the ablation study on the \textit{Improper dataset pair} case for three types of datasets (\#1, indoor flat floor; \#2, wood tiles; and \#6, grass in Figure~\ref{fig:dataset}). Note that the results of the proposed method (Ours) in this table are the same as those in Table~\ref{tab:ablation_study_of_identicalMLP3}. While the \textit{Improper dataset pair} case cannot express features fitting to proper terrain conditions, our network was trained to adapt to the current environment. Hence, this result shows that our network was trained properly.}
  \label{tab:ablation_study_of_Improper_dataset_pair}
  \begin{threeparttable}
    \setlength{\tabcolsep}{1.7mm}
    \begin{tabular}{cc|cc|cc|cc}
      \hline
      \multicolumn{2}{c|}{\multirow{-0.5}{*}{Method/Seq.}} & \multicolumn{2}{c|}{\#1} & \multicolumn{2}{c|}{\#2} & \multicolumn{2}{c}{\#6} \\ \cline{3-8}
      
      \multicolumn{2}{c|}{\multirow{-2}{*}{}} & $t_{\rm{RTE}}$ & \cellcolor[HTML]{EFEFEF}$r_{\rm{RTE}}$ & $t_{\rm{RTE}}$ & \cellcolor[HTML]{EFEFEF}$r_{\rm{RTE}}$ & $t_{\rm{RTE}}$ & \cellcolor[HTML]{EFEFEF}$r_{\rm{RTE}}$ \\ \hline \hline
      
      \multicolumn{2}{c|}{Ours} & \textbf{0.017} & \cellcolor[HTML]{EFEFEF}{0.24} & \textbf{0.016} & \cellcolor[HTML]{EFEFEF}{0.25} & \textbf{0.038} & \cellcolor[HTML]{EFEFEF}{0.31} \\ 

       \multicolumn{2}{c|}{Case 1\tnote{1}} & 0.033 & \cellcolor[HTML]{EFEFEF}0.36 & 0.034 & \cellcolor[HTML]{EFEFEF}0.24 &\slashbox[10mm]{}{} & \cellcolor[HTML]{EFEFEF}\slashbox[10mm]{}{} \\
       
      \multicolumn{2}{c|}{Case 2\tnote{2}} & 0.15 & \cellcolor[HTML]{EFEFEF}0.36 & \slashbox[10mm]{}{} & \cellcolor[HTML]{EFEFEF}\slashbox[10mm]{}{} & 0.165 & \cellcolor[HTML]{EFEFEF}0.34 \\ \hline

    \end{tabular}
    \begin{tablenotes}[flushleft]
      \item[1] Training: indoor flat floor; Testing: wood tiles.
      \item[2] Training: indoor flat floor; Testing: grass.
    \end{tablenotes}  \end{threeparttable}
\end{table}
\begin{table*}[]
  \caption{Ratio of frames facing degeneration and absence of the point clouds. Especially, \textit{Seq.~1} was a difficult condition because the majority of point clouds (91~\%) were degenerated or unavailable.}
  \label{tab:points_states}  
  \begin{tabular}{cl|cc|cc|c}
  \hline
  \multicolumn{2}{c|}{}                             & \multicolumn{2}{c|}{\textbf{Point cloud degeneracy}}                       & \multicolumn{2}{c|}{\textbf{Point cloud absence}}                           \\ \cline{3-6} 
  \multicolumn{2}{c|}{\multirow{-2}{*}{Sequences/Problems}} & Ratio to all frames  & \cellcolor[HTML]{EFEFEF}Duration {[}s{]} & Ratio to all frames                           & \cellcolor[HTML]{EFEFEF}Duration {[}s{]} & Total time {[}s{]} \\ \hline \hline
  \multicolumn{2}{c|}{Seq.~1} & 69~\% (572$/$828) & \cellcolor[HTML]{EFEFEF}57.2  & 23~\% (189$/$828) & \cellcolor[HTML]{EFEFEF}18.9   & 85.7      \\ 
  \multicolumn{2}{c|}{Seq.~2}                  & 26~\% (110$/$429)          & \cellcolor[HTML]{EFEFEF}11.1             & 2~\% (9$/$429)             & \cellcolor[HTML]{EFEFEF}0.9       & 76.3         \\ 
  \multicolumn{2}{c|}{Seq.~3}                  & 30~\% (227$/$760)          & \cellcolor[HTML]{EFEFEF}22.7             & 8~\% (60$/$760)            & \cellcolor[HTML]{EFEFEF}6.0     & 41.8              \\ \hline
  \end{tabular}
\end{table*}

We used a Rover mini (Rover Robotics) equipped with two types of LiDAR, as shown in Figure~\ref{fig:skid_steering_robot}.
To conduct the offline learning procedure, we recorded the point cloud and IMU data using omnidirectional FOV LiDAR (Livox Mid-360), and the wheel angular velocities using each wheel encoder.
Specifically, point clouds, IMU measurements, and the wheel angular velocities were recorded at 10, 200, and 60 Hz, respectively. 
In this experiment, IMU measurements and wheel angular velocities were used as input data for the proposed neural network inferring the 2D twist of a skid-steering robot. 
Whereas, point clouds and IMU measurements were used for creating reference data of the 2D twist based on tightly coupled LiDAR-IMU odometry in feature-rich environments.
We used $70~\%$ and $30~\%$ of the dataset (Figure~\ref{fig:dataset}) for training and validation, respectively.

Table~\ref{tab:ablation_study_of_identicalMLP3} shows the relative trajectory errors (RTEs) for the proposed method (Ours) and \textit{Identical network}.
The translation RTEs of the proposed method~(Mean: \SI{0.021}{m}) were about 1.5 times as accurate as those of the \textit{Identical network}~ (Mean: \SI{0.031}{m}) in all datasets because translational accuracy depends on terrain condition changes. 
The difference in rotational accuracy between these methods was slight (\SI{0.02}{deg}).
This result shows the proposed network can properly adapt to each terrain condition.

Table~\ref{tab:ablation_study_of_Improper_dataset_pair} presents the RTEs of the proposed method and the \textit{Improper dataset pair} case for three types of datasets (indoor flat floor, wood tiles, and grass). 
This result also shows that the proposed method (Ours) is the most accurate translation RTEs in all cases.
The average of translation RTEs of our method in terrains \#1 and \#2~(\SI{0.017}{m}) was twice as accurate as that of Case~1~(\SI{0.034}{m}).
In addition, the average of translation RTEs of our method in terrains \#1 and \#6~(\SI{0.028}{m}) was about six times as accurate as that of Case~2~(\SI{0.158}{m}).
Whereas, the accuracy of rotation RTEs is almost the same in all cases.
Furthermore, the translation RTEs of Case 1 are more accurate than those of Case 2 because, while the difference between \textit{indoor flat floor} and \textit{wood tiles} (Case~1) is slight, the difference between \textit{indoor flat floor} and \textit{grass} (Case~2) is significant. 
Hence, this result implies that the network trained for an \textit{indoor flat floor} was almost the same as the network trained for \textit{wood tiles} compared with another network trained for \textit{grass}.
Clearly, \textit{indoor flat floor} was more similar to another flat and hard terrain (\textit{wood tiles}) than to grass (i.e., rough and soft terrain).
Therefore, this result demonstrates that the proposed network was trained to properly represent features that vary with the environment.

\subsection{Accuracy evaluation of the proposed odometry estimation algorithm}\label{subsection:LIWOresults}
We conducted experiments to evaluate the robustness of the proposed method against severe point cloud degeneration under conditions producing large wheel slippage (i.e., nonlinear kinematic behavior of the skid-steering robot).
We also evaluated adaptability of our network to changing terrain conditions.
We operated the robot testbed (Figure~\ref{fig:skid_steering_robot}) at a maximum speed of \SI{3}{m/s} (the maximum rotational speed of each wheel is \SI{36}{rad/s}) to induce significant wheel slippage. 
Refer to Video~1~\footnote{\url{https://youtu.be/ZkGO21Q_paY}} for an understanding of the maximum wheel speed as a reference.
In contrast to the evaluation described in Section~\ref{subsection:FoooNETresults}, we applied the narrow FOV LiDAR (Livox AVIA) to the proposed odometry estimation algorithm to imitate severe point cloud degeneration in, for example, tunnels.
Furthermore, we placed the narrow FOV LiDAR toward the Y direction (side) of the robot frame (Figure~\ref{fig:wheel_robot_description}).
Hence, degenerated point clouds were obtained when the robot moved near a wall as shown in Figure~\ref{fig:estimated_traj}-(d).
In this experiment, we input all wheel encoder values, IMU measurements, and the point cloud measured by the narrow FOV LiDAR into the proposed odometry estimation algorithm for estimating the robot state ${\bm X}_t$ and MLP parameters ${\bm P}_t$ on a unified factor graph, as shown in Figure~\ref{fig:overview}.
\begin{figure}[tb]
  \centering
  \includegraphics[width=0.6\linewidth]{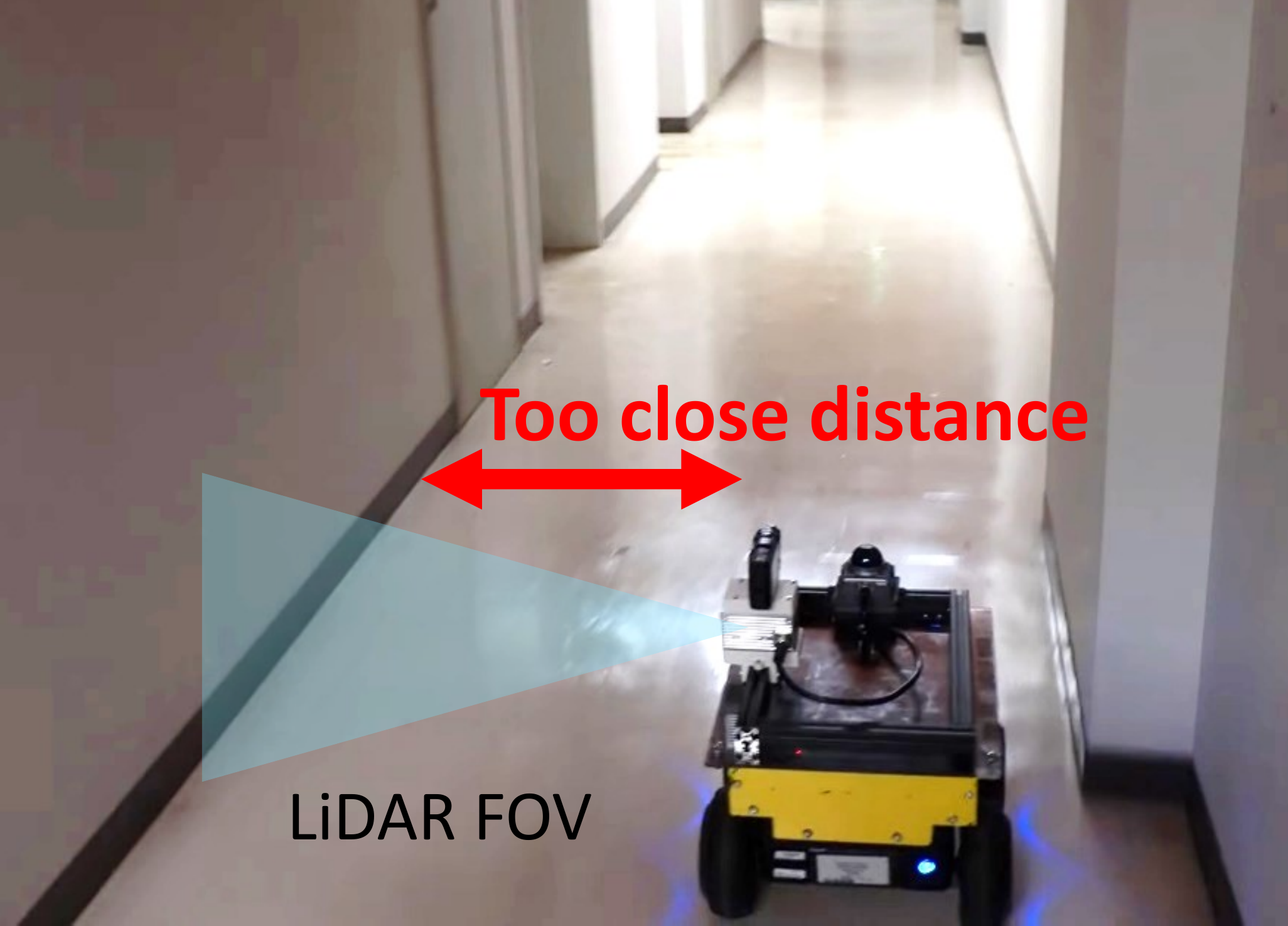}
  \caption{Example of situations where point clouds are unavailable due to the distance between the LiDAR and walls being too small. For the Livox AVIA LiDAR used for odometry estimation in this experiment, the minimum observation range is \SI{1}{m}. Thus, we also verified the robustness of our odometry estimation algorithm to point cloud absence in addition to point cloud degeneration.
           }
  \label{fig:too_close}
\end{figure}

The proposed method was evaluated for three sequence types. \textit{Seq.~1} included severe degeneracy of point clouds in straight long corridors and drifting (Figure~\ref{fig:estimated_traj}-(e)). 
In the case of \textit{Seq.~2}, the robot transited flat hard terrain, such as bricks, outdoor stone tiles, and indoor stone tiles, as shown in Figure~\ref{fig:estimated_traj}-(b). 
\textit{Seq.~3} included drastic terrain changes from concrete to grass, as shown in Figure~\ref{fig:estimated_traj}-(c). 

Table~\ref{tab:points_states} summarizes the number of frames where the point cloud data became 1) degenerate or 2) unavailable (fewer than 100 points in a frame). 
The narrow FOV LiDAR was always pointed at the walls in all sequences; thus, the observed point clouds were mostly degenerated. 
Furthermore, the point cloud measurements became completely unavailable when the sensor approached the wall (Figure~\ref{fig:too_close}) because the LiDAR cannot observe points closer than the minimum observation range (\SI{1}{m}). 
\textit{Seq.~1} was an especially difficult condition compared with the other sequences because the majority of point clouds (91~\%) were degenerated or unavailable.

Note that the omnidirectional FOV LiDAR (Livox MID-360) and its IMU data were used only for estimating the reference trajectory as ground truth.
To obtain the reference trajectory, we performed a batch optimization of LiDAR-IMU constraints on a precise prior map.
The prior map was built by a 3D laser scanner (FARO Focus3D X330).
More specifically, to estimate the sensor trajectory, we minimize the following objective function \mbox{$e(\chi)^{\text{gt}}$} for all robot states \mbox{$\chi$} (i.e., ground truth).
\begin{equation}
  \label{eq:gt_estimation}
  \begin{aligned}
    e(\chi)^\text{gt} &= 
    \underbrace{\sum_{x_i \in \chi} \sum_{j=i-N^{\mathrm{fc}}}^{i-1} e^{\mathrm{M}}\left(\mathcal{P}_i, \mathcal{P}_j, \boldsymbol{T}_i, \boldsymbol{T}_j\right)}_{\text{Scan-to-scan matching cost factor}} \\
    &\quad + \underbrace{\sum_{x_i \in \mathcal{X}} e^{\mathrm{M}}\left(\mathcal{P}_i, \mathcal{M}, \boldsymbol{T}_i, \mathbb{I}^{4 \times 4}\right)}_{\text{Scan-to-map matching cost factor}} \\
    &\quad + \underbrace{\sum_{x_i \in \mathcal{X}} e^{\mathrm{IMU}}\left(\boldsymbol{x}_i, \boldsymbol{x}_j\right)}_{\text{IMU preintegration factor}}
  \end{aligned}
\end{equation}
The objective function is defined by the scan-to-scan-based and the scan-to-map matching cost factors, and IMU preintegration factors for all robot states \mbox{$\chi$}.
Here, \mbox{$N^{\text{fc}}$} is the number of fully connected LiDAR frames, \mbox{$\mathcal{M}$} is a point cloud of the prior map, and \mbox{$\mathbb{I}^{4\times 4}$} is the 4-by-4 identity matrix.
We carefully inspected the estimated trajectory to ensure that all point clouds of the omnidirectional LiDAR are precisely aligned with the prior map.
To solve this optimization problem, we used the LiDAR-IMU localization~\mbox{\cite{koide2024glil}} results with the omnidirectional LiDAR as the initial values of the robot states.
We used this prior map-based method to obtain ground truth of trajectories because this approach can be applied to all our sequences in contrast to other methods such as a motion tracking system, total station, or Global Navigation Satellite System (GNSS).
For example, in the case of Seq.2 including the transition from outdoor to indoor environments, it is difficult for the aforementioned three methods to obtain their measurements properly.

We set initial values of ${\bm P}_t$ (weights and biases of the \textit{online learning model}) by using the corresponding values obtained via the offline learning procedure.
In our implementation, we set these parameters trained for the indoor flat floor (i.e., \#1 in Figure~\ref{fig:dataset}) as the initial value of ${\bm P}_1$.
We consider that the indoor flat floor case is neutral for a kinematic model of wheel robots compared to other terrains (e.g., grass) in terms of sufficient flatness and friction.
We chose the parameters trained for the indoor flat floor (i.e., \mbox{$\boldsymbol{M}_{\text {on}, \# 1}$} in the offline learning phase) as the initial value \mbox{${\bm P}_1$} because we expected that this parameter could be adapted to various terrain conditions compared to others.

\begin{table*}[t]
  \centering
  \caption{ATEs and RTEs of the odometry estimation algorithms. The proposed method (\textit{Ours}) was twice as accurate as \textit{Ours w/o online learning} owing to online training. 
  \textit{LIO w/ online calibrated full linear model} and \textit{FAST-LIO-Multi-Sensor-Fusion} suffered from complex robot motions (e.g., large wheel slippage) because these methods regarded the kinematic model as linear. 
  \textit{FAST-LIO2} failed in all sequences because all environments included locations where point clouds degenerate.}

  \label{tab:ATE_and_RTE}
  \begin{threeparttable}
    \setlength{\tabcolsep}{1.11mm} 
    \begin{tabular}{cc|cc|cc|cc}
      \hline
      \multicolumn{2}{c|}{\multirow{-0.5}{*}{Method/Sequence}} & \multicolumn{2}{c|}{Seq.~1} & \multicolumn{2}{c|}{Seq.~2} & \multicolumn{2}{c}{Seq.3} \\ \cline{3-8}
      
      \multicolumn{2}{c|}{\multirow{-2}{*}{}} & ATE~[\SI{}{m}] & \cellcolor[HTML]{EFEFEF}RTE~[\SI{}{m}] & ATE~[\SI{}{m}] & \cellcolor[HTML]{EFEFEF}RTE~[\SI{}{m}] & ATE~[\SI{}{m}] & \cellcolor[HTML]{EFEFEF}RTE~[\SI{}{m}] \\ \hline \hline
      
      \multicolumn{2}{c|}{Ours} & 
      \textbf{1.07} $\pm$ 0.35 &     \cellcolor[HTML]{EFEFEF}{\textbf{0.08}}  $\pm$ 0.05 & 
      \textbf{0.25} $\pm$ 0.10 &     \cellcolor[HTML]{EFEFEF}{\textbf{0.12}}  $\pm$ 0.07 & 
      \textbf{0.41} $\pm$ 0.18 &     \cellcolor[HTML]{EFEFEF}{\textbf{0.09}}  $\pm$ 0.07 \\ 
      
      \multicolumn{2}{c|}{Ours w/o online learning} & 
      3.29 $\pm$ 2.42 & \cellcolor[HTML]{EFEFEF}{0.09}  $\pm$ 0.06 & 
      0.59 $\pm$ 0.53 & \cellcolor[HTML]{EFEFEF}{0.16}  $\pm$ 0.07 & 
      0.99 $\pm$ 0.51 & \cellcolor[HTML]{EFEFEF}{0.11}  $\pm$ 0.07 \\ 

      \multicolumn{2}{c|}{LIO\,w/\,online\,calibrated\,full\,linear\,model\,\cite{okawara2024tightly}} &   
      18.80 $\pm$ 14.40 & \cellcolor[HTML]{EFEFEF}{0.08}  $\pm$ 0.07 & 
      2.23 $\pm$ 2.16 & \cellcolor[HTML]{EFEFEF}{0.12}  $\pm$ 0.08 & 
      0.98 $\pm$ 0.50 & \cellcolor[HTML]{EFEFEF}{0.09}  $\pm$ 0.05 \\

      \multicolumn{2}{c|}{FAST-LIO-Multi-Sensor-Fusion~\mbox{\cite{10272296}}} &       
      7.99 $\pm$ 3.10 & \cellcolor[HTML]{EFEFEF}{0.36}  $\pm$ 0.31 & 
      2.41 $\pm$ 2.17 & \cellcolor[HTML]{EFEFEF}{0.49}  $\pm$ 0.46 & 
      2.49 $\pm$ 1.63 & \cellcolor[HTML]{EFEFEF}{0.17}  $\pm$ 0.22 \\
      
      \multicolumn{2}{c|}{FAST-LIO2~\cite{xu2022fast}} & 
      Error $>$\SI{20}{m} & \cellcolor[HTML]{EFEFEF}{Error $>$\SI{20}{m}} & 
      Error $>$\SI{20}{m} & \cellcolor[HTML]{EFEFEF}{Error $>$\SI{20}{m}} & 
      Error $>$\SI{20}{m} & \cellcolor[HTML]{EFEFEF}{Error $>$\SI{20}{m}}  \\

      \hline

    \end{tabular}

  \end{threeparttable}
\end{table*}

\begin{table*}[tb]
  \caption{Comparison of total processing times regarding odometry estimation including point cloud preprocessing and IMU preintegration as well as online learning (factor graph optimization). All methods perform odometry estimation at sufficiently high speeds, which are sufficiently faster than the frequency of LiDAR point cloud updates (\mbox{\SI{0.1}{s}}).}
  \label{tab:time_comparison}
  \centering
  \scriptsize
  \small
  \begin{tabular}{lc}
  \toprule
  Methods & Total processing time [s]\\
  \midrule
  Ours & 0.031 \\
  Ours w/o online learning & 0.017 \\
  LIO w/ online calibrated full linear model~\mbox{\cite{okawara2024tightly}} & 0.023 \\
  FAST-LIO-Multi-Sensor-Fusion~\mbox{\cite{10272296}} & 0.007 \\
  FAST-LIO2~\mbox{\cite{xu2022fast}} & 0.005 \\
  \bottomrule
  \end{tabular}
\end{table*}

\subsubsection{Comparison with state-of-the-art methods}\label{subsubsec:odometry_evaluation}
We compared the proposed odometry estimation algorithm (\textit{Ours}) with the following methods:
\begin{itemize}
  \item \textit{Ours w/o online learning}: We used the fixed neural network (i.e., the \textit{Identical network} case) described in Section~\ref{subsection:FoooNETresults} instead of the proposed network for conducting an ablation study. We expected that the fixed network cannot adapt to a proper terrain condition.
  \item \textit{LIO w/ online calibrated full linear model} (our previous work)~\mbox{\cite{okawara2024tightly}}: This algorithm fuses matching cost factors (Section~\mbox{\ref{subsec:MCF}}) and IMU preintegration factors (Section~\mbox{\ref{subsec:IPF}}), and the full linear model~\mbox{\cite{anousaki2004dead}}-based constraints for odometry estimation.
  This algorithm also conducts online calibration of the full linear model~\mbox{\cite{anousaki2004dead}} to adapt to terrain condition changes and kinematic parameter errors.
  The difference between the proposed method and our previous work is the type of the kinematic model-based constraints.
  While the proposed method incorporates the neural network-based kinematic model as a constraint, our previous work utilizes the full linear model-based kinematic model; therefore, we expected that the full linear model cannot express nonlinearity (e.g., large wheel slippage).
  \item \textit{FAST-LIO2}~\cite{xu2022fast}: FAST-LIO2 is the state-of-the-art LiDAR-IMU odometry based on tightly-coupled LiDAR and IMU fusion, as described in Section~\ref{subsec:LIO}.
  \item \href{https://github.com/zhh2005757/FAST-LIO-Multi-Sensor-Fusion}{\it{FAST-LIO-Multi-Sensor-Fusion}}~\cite{10272296}: Zhao et~al. fused wheel odometry-based constraints~\cite{10272296} and FAST-LIO2~\cite{xu2022fast} to improve estimation performance. They considered only longitudinal motion in the kinematic model, without taking into account lateral and rotational movements, even for a skid-steering robot. Furthermore, the wheel odometry-based constraint was directly defined by the longitudinal velocity of the robot, not the kinematic model (i.e., the wheel odometry-based constraint was incorporated by the loose coupling way).
\end{itemize}

\begin{figure*}[tb]
  \centering
  \includegraphics[width=1\linewidth]{figs/estimated_traj_gray.pdf}
  \caption{Odometry estimation results: (a), (b), and (c) are trajectory comparison results for Seq.~1, Seq.~2, and Seq.~3, respectively. (See Section 4.2 for additional discussion on the evaluation sequences.)
  The dotted black ellipses indicate areas where point clouds degenerate.
  The narrow FOV LiDAR was pointed at straight walls in all sequences to validate the robustness to severe point cloud degeneration.
  (d) point clouds lacking geometric features were observed when the LiDAR was pointed only at straight walls (d-1, d-2, and d-3).
  (e) Seq.~1 included complex robot motions involving large wheel slippage (e.g., drifting) in long straight corridors (d-1) where LiDAR point clouds degenerate. 
  In Seq.~2, the robot transited three types of flat hard terrain and faced point cloud degeneration two times~(d-2).
  Furthermore, in the case of Seq.~3, the robot transited from concrete to grass (i.e., rough and soft terrain) and also faced point cloud degeneration on the grass~(d-3). The aerial photograph in (c) was obtained by Google Maps.
  In contrast to our previous work~\mbox{\cite{okawara2024tightly}}, we operated the robot testbed at the maximum rotational speed of each wheel \mbox{\SI{36}{rad/s}} (i.e., nine times faster than that of our previous work~\mbox{\cite{okawara2024tightly}}) to induce significant wheel slippage (Video~1).
 }
  \label{fig:estimated_traj}
\end{figure*}

Table~\ref{tab:ATE_and_RTE} shows the absolute trajectory errors (ATEs) and relative trajectory errors (RTEs)~\cite{zhang2018tutorial} for all methods.
We also show the trajectory comparison results in Figure~\ref{fig:estimated_traj}.
According to these results, in terms of ATEs and RTEs, the proposed method (\textit{Ours}) outperformed all other methods in all sequences owing to the neural adaptive odometry factor, which can capture the nonlinearity of the kinematic model to adapt it to the current terrain condition.
\textit{FAST-LIO2} failed due to point cloud degeneration when the LiDAR pointed only at the wall in all sequences.
\textit{LIO w/ online calibrated full linear model} and \textit{FAST-LIO-Multi-Sensor-Fusion} suffered from nonlinearities such as drifting (executed at each corner of corridors in Figure~\ref{fig:estimated_traj}-(a)) and successive left and right turning in rough terrain (executed in the grass, Figure~\ref{fig:estimated_traj}-(c)) because these methods create constraints based on the linear kinematic model of the skid-steering robot.
\textit{Ours w/o online learning} approximately captured the nonlinearities compared with other comparative methods.
However, \textit{Ours w/o online learning} could not adapt to changes in terrain, and thus scales of the estimated trajectories were not consistent (especially, Figure~\ref{fig:estimated_traj}-(b), (c)) in contrast to the proposed method with online learning. 
The ATEs of our method (Seq.~1: \SI{1.07}{m}; Seq.~2: \SI{0.25}{m}; Seq.~3: \SI{0.41}{m}) were more than twice as accurate as a method incorporating a fixed network (i.e., batch learning rather than online learning)-based constraints (Seq.~1: \SI{3.29}{m}; Seq.~2: \SI{0.59}{m}; Seq.~3: \SI{0.99}{m}) in all sequences.

We confirmed that the processing times of online training for all sequences were approximately \mbox{\SI{0.01}{s}} on a laptop PC equipped with the NVIDIA GeForce RTX 3060 Laptop GPU and the AMD Ryzen 9 5900HX processor.
Table~\mbox{\ref{tab:time_comparison}} presents total processing times of all methods, including point cloud preprocessing and IMU preintegration as well as online learning (factor graph optimization).  
The proposed method requires a longer processing time compared to those of other methods due to the computational cost of training of the neural network.  
While our online learning method optimizes MLP parameters \mbox{$\bm {P}_i~\in~\mathbb{R}^{100}$} online, our previous method~\mbox{\cite{okawara2024tightly}} optimizes the kinematic parameters \mbox{$\bm {K}_i~\in~\mathbb{R}^{6}$} online.  
However, the processing times of online learning (\mbox{\SI{0.01}{s}}) and the entire process (\mbox{\SI{0.03}{s}}) are sufficiently faster than the frequency of LiDAR point cloud updates (\mbox{\SI{0.1}{s}}).  
Therefore, we consider that the proposed method can meet real-time constraints in most wheel robot applications.

In addition, Video~2~\footnote{\url{https://youtu.be/kTKAsCFPhOg}}, Video~3~\footnote{\url{https://youtu.be/YU-MZFODQkg}}, and Video~4~\footnote{\url{https://youtu.be/WdSyUgZa6jA}} show supplementary material for understanding each terrain's characteristics and the robot's speed for Seq.~1, Seq.~2, and Seq.~3, respectively.

\begin{figure}[tb]
  \centering
  \includegraphics[width=1\linewidth]{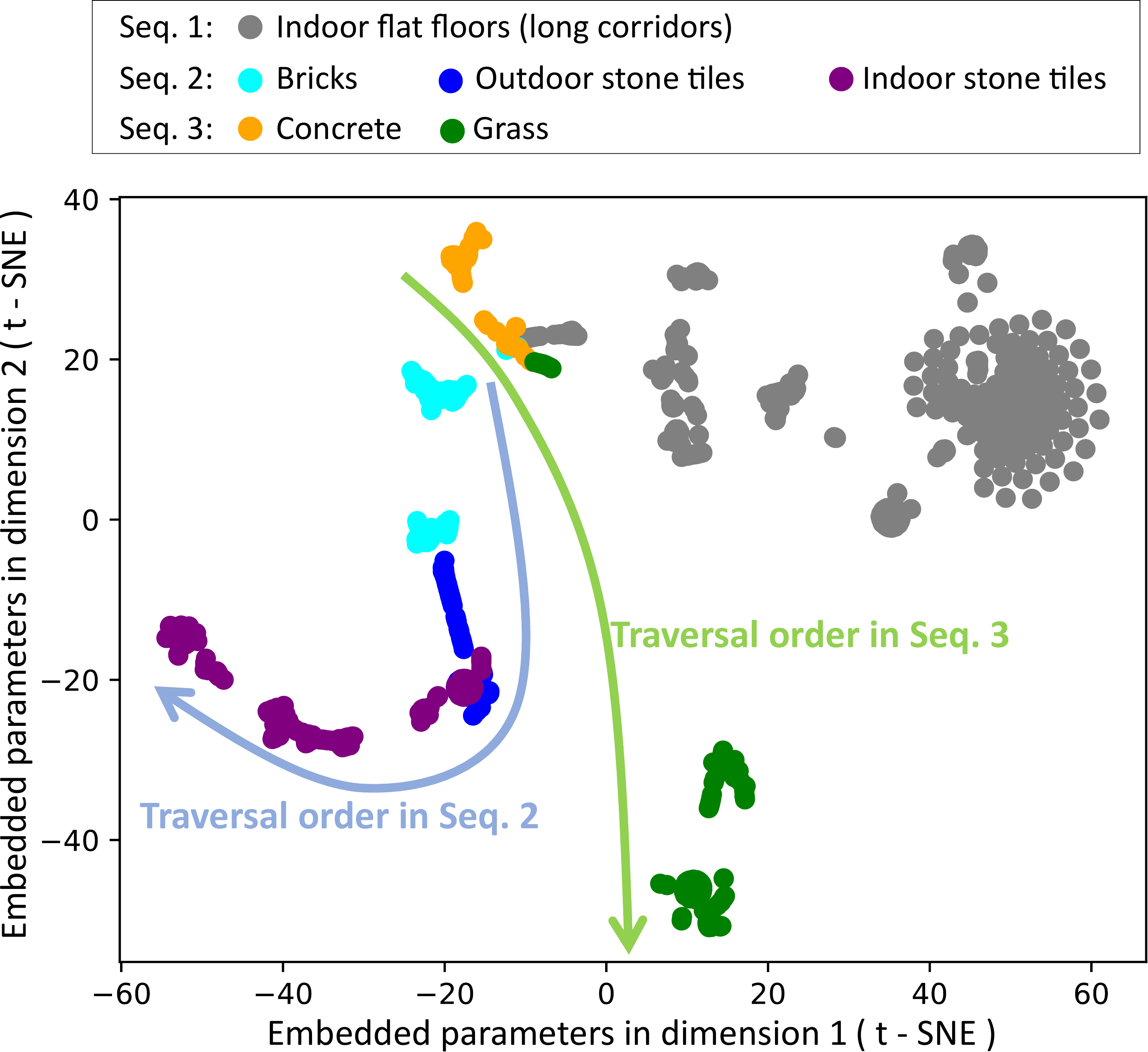}
  \caption{Visualization of the time histories of MLP parameters $\bm P$ (100 dimensions) for all sequences with 2D t-SNE embedding.
  In \textit{Seq.~1}, the robot moved on similar terrains consistently.
  \textit{Seq.~2} included terrain changes from bricks to outdoor stone tiles to indoor stone tiles. 
  \textit{Seq.~3} also included changes in terrain conditions from asphalt to grass.
  The embedded parameters of grass (uneven and soft) were isolated from other terrains (flat and hard).
  Both Seq.~3 and Seq.~2 included terrain condition changes; however, the embedded parameters in Seq.~3 changed significantly compared to those in Seq.~2.
           }
  \label{fig:tsne2d}
\end{figure}
\subsubsection{Adaptability evaluation of the proposed network with online learning}\label{Adaptability}
To verify adaptability of the proposed network online trained by the \textit{neural adaptive odometry factor} with LiDAR-IMU-wheel odometry, we visualized the trained weight and bias of neurons in each sequence.
We summarized those values by using t-distributed Stochastic Neighbor Embedding (t-SNE)~\cite{van2008visualizing}, which embeds high-dimensional data into low-dimensional data through manifold learning.
Figure~\ref{fig:tsne2d} shows the embedded parameters (2 dimensions) from time histories of MLP parameters $P$ (100 dimensions) for all sequences.
We can see that the embedded parameters for grass are isolated from the parameters of other terrains.
This result is reasonable because the grass was uneven and soft terrain that was distinct from other flat and hard terrains.
In the case of \textit{Seq.~3} (transition from concrete to grass), the embedded parameters drastically changed during this transition because this difference between these terrains was significant.
In contrast, in  \textit{Seq.~2} (transition from bricks to outdoor stone tiles to indoor stone tiles), the embedded parameters gradually changed because the differences between these terrains were slight.
Similarly, in \textit{Seq.~1} (indoor flat floors as in long corridors), the embedded parameters' behavior differed from those in each terrain to describe the proper terrain-dependent features.

Accordingly, these results indicate that the proposed network can adapt to the current terrain condition owing to online learning by \textit{neural adaptive odometry factor}.

\section{Conclusion}
We proposed the LiDAR-IMU-wheel odometry algorithm incorporating online training of the neural network to infer the kinematic model of the skid-steering robot.
We designed the proposed network to include an \textit{offline learning model} and \textit{online learning model}~(Figure~\ref{fig:FoooNET}).
The \textit{offline learning model} was trained offline with datasets for eight types of terrain (Figure~\ref{fig:dataset}) to learn features related to terrain-independent parameters~(Figure~\ref{fig:offline_and_online}).
In contrast, the \textit{online learning model} was trained via the \textit{neural adaptive odometry factor} with LiDAR-IMU-wheel odometry on a unified factor graph to ensure consistency with all those constraints.

We showed that the proposed network was more accurate than other networks not properly adapted to each terrain condition owing to our network design.
We compared the proposed odometry estimation algorithm with the state-of-the-art methods~\cite{xu2022fast, 10272296, okawara2024tightly}.
The proposed method outperformed those methods owing to neural network-based constraints, which can express the complicated nonlinearity (e.g., large wheel slippage such as drifting) of the kinematic model of a skid-steering robot.

In future work, we plan to improve versatility of our network to adapt to any type of wheeled robot model (e.g., different size and mechanism from those used during the offline training phase) in addition to terrain condition changes. We need to enlarge the size of the network to enhance its expression by implementing the online adaptive odometry factor (i.e., online learning) with GPU acceleration, as in~\cite{koide2021globally}.



\bibliographystyle{cas-model2-names}


\bibliography{cas-refs}

\end{document}